\documentclass{elsarticle}

\usepackage[utf8]{inputenc}
\newcommand{\argmax}{\operatorname*{\arg\max}}

\usepackage[colorlinks=true,allcolors=black,breaklinks=true]{hyperref}
\usepackage{multicol}
\usepackage{multirow}
\usepackage{amsmath}
\usepackage{amssymb}
\usepackage{natbib}
\usepackage{tabularx}
\usepackage{booktabs}
\usepackage{setspace}
\usepackage{algpseudocode}
\usepackage{hyperref}
\usepackage{subfig}
\usepackage{tikz}
\usetikzlibrary{arrows,automata,backgrounds,fit,patterns,petri,shadows,shapes}

\pgfdeclarepatternformonly{stripes}{\pgfpoint{0cm}{0cm}}{\pgfpoint{1cm}{1cm}}{\pgfpoint{1cm}{1cm}}{
  \foreach \i in {0.125, 0.375,...,0.875}{
    \pgfpathmoveto{\pgfpoint{\i cm}{0cm}}
    \pgfpathlineto{\pgfpoint{1cm}{1cm - \i cm}}
    \pgfpathlineto{\pgfpoint{1cm}{1cm - \i cm + 0.125cm}}
    \pgfpathlineto{\pgfpoint{\i cm - 0.125cm}{0cm}}
    \pgfpathclose
    \pgfusepath{fill}
    \pgfpathmoveto{\pgfpoint{0cm}{\i cm}}
    \pgfpathlineto{\pgfpoint{1cm - \i cm}{1cm}}
    \pgfpathlineto{\pgfpoint{1cm - \i cm - 0.125cm}{1cm}}
    \pgfpathlineto{\pgfpoint{0cm}{\i cm + 0.125cm}}
    \pgfpathclose
    \pgfusepath{fill}
  }
}
\tikzstyle{decisionA} = [regular polygon, regular polygon sides = 4, thick, minimum size = 1.25cm, inner sep = 0.1pt, draw = black, fill = gray!40]
\tikzstyle{decisionC} = [regular polygon, regular polygon sides = 4, thick, minimum size = 1.25cm, inner sep = 0.1pt, draw = black]
\tikzstyle{utilityA} = [regular polygon, regular polygon sides = 6, thick, minimum size = 1cm, inner sep = 0.1pt, draw = black, fill = gray!40]
\tikzstyle{utilityC} = [regular polygon, regular polygon sides = 6, thick, minimum size = 1cm, inner sep = 0.1pt, draw = black]
\tikzstyle{chanceA} = [circle, thick, minimum size = 1cm, inner sep = 0.1pt, draw = black, fill = gray!40]
\tikzstyle{chanceC} = [circle, thick, minimum size = 1cm, inner sep = 0.1pt, draw = black]
\tikzstyle{chance} = [circle, thick, minimum size = 1cm, inner sep = 0.1pt, draw = black, pattern = stripes, pattern color = gray!40]
\tikzstyle{empty} = [circle, line width = 0pt, minimum size = 1cm, inner sep = 0.1pt]
\biboptions{authoryear}
\usepackage{lineno,hyperref}
\modulolinenumbers[5]
\journal{International Journal of Approximate Reasoning}

\begin{document}

\begin{frontmatter}
\title{Adversarial classification: An adversarial risk analysis approach}

\author[mymainaddress]{Roi Naveiro\corref{mycorrespondingauthor}}
\cortext[mycorrespondingauthor]{Corresponding author}
\ead{roi.naveiro@icmat.es}

\author[mymainaddress]{Alberto Redondo}
\ead{alberto.redondo@icmat.es}

\author[mymainaddress]{David Ríos Insua}
\ead{david.rios@icmat.es}

\author[mysecondaddress]{Fabrizio Ruggeri}
\ead{fabrizio@mi.imati.cnr.it}

\address[mymainaddress]{Institute of Mathematical Sciences (ICMAT-CSIC) Campus Cantoblanco UAM, C/ Nicolás Cabrera, 13-15, 28049 Madrid, Spain.}

\address[mysecondaddress]{CNR-IMATI, Via Alfonso Corti, 12, 20133, Milano, Milano, Italy}

\begin{abstract}
Classification techniques are widely used in security settings in which data can be deliberately manipulated by an adversary trying to evade detection and achieve some benefit. However, traditional classification systems are not robust to such data modifications. Most attempts to enhance classification algorithms in adversarial environments have focused on game theoretical ideas under strong underlying common knowledge assumptions, which are not actually realistic in security domains. We provide an alternative framework to such problems based on adversarial risk analysis which we illustrate with examples. Computational, implementation and robustness issues are discussed.
\end{abstract}
%
\end{frontmatter}
\section{Introduction}
\label{sec:introduction}
Classification is one of the most widely used instances of supervised learning, with applications in areas such as bioinformatics, \cite{bioinf}; spam detection, \cite{spam}; credit scoring, \cite{scoring}; computer vision, \cite{chen2015handbook}; and genomics, \cite{genom}. In recent years, the field has experienced an enormous growth becoming a major research area in statistics and machine learning, \cite{efron2016computer}. Most efforts in classification have focused on obtaining more accurate algorithms which, however, largely ignore a relevant issue in many application areas: the presence of adversaries who can actively manipulate data to fool the classifier so as to attain a benefit.

As a motivating example consider the case of fraud detection. As machine learning algorithms are incorporated to such detection task, fraudsters begin to learn how to evade them. For instance, they could find out that making a huge transaction increases the probability of being detected and start issuing smaller transactions more frequently rather than a single big one. The presence of adaptive adversaries has been pointed out in areas such as spam detection, \cite{zeager2017adversarial}; fraud detection, \cite{Kocz2009FeatureWF}; and computer vision, \cite{goodfellow}. In such contexts, algorithms should take into account possible modifications on the behaviour of adversaries so as to be robust against adversarial data manipulations. 

\cite{adversarialClassification2004} provided a pioneering approach to enhance classification algorithms when an adversary is present, calling it adversarial classification (AC). They view AC as a game between a classifier $C$ and an adversary $A$. The classifier aims at finding an optimal classification strategy against $A$'s optimal attacking strategy. Computing Nash equilibria, \cite{ozdaglar2011network}, in such general games becomes overly complex. Therefore, they propose a simplified version in which \textit{C} first assumes that data is untainted and computes her optimal classifier; then, \textit{A} deploys his optimal attack against it; subsequently, \textit{C} implements the optimal classifier against this attack, and so on. As the authors pointed out, a very strong assumption is made: all parameters of both players are known to each other. Although standard in game theory, this common knowledge assumption is actually unrealistic in security scenarios.

Stemming from this work, there has been an important literature in AC, reviewed in \cite{biggio2014security} or \cite{li2014feature}. Subsequent approaches have focused on analyzing attacks over classification algorithms and assessing their robustness against such attacks. 
To that end, some assumptions about the adversary are made. For instance, \cite{adversarialLearning2005} consider that the adversary is able to send membership queries to the classifier, the entire feature space being known to issue optimal attacks; then, they prove the vulnerability of linear classifiers against adversaries. Similarly, \cite{zhou2012adversarial} consider that the adversary seeks to push his malicious instances into innocuous ones, assuming that the adversary can estimate such instances. 

A few methods have been proposed to robustify classification algorithms in adversarial environments. Most of them have focused on application-specific domains, as \cite{Kocz2009FeatureWF} on spam detection. \cite{Vorobeychik:2014:ORC:2615731.2615811} study the impact of randomization schemes over different classifiers against adversarial attacks proposing an optimal randomization scheme as best defense. Other approaches have focused on improving the game theoretic model in \cite{adversarialClassification2004} but, to our knowledge, none has been able to overcome the unrealistic common knowledge assumptions, as may be seen in recent reviews by \cite{biggio2018wild} and \cite{surveygametheoretic}, which have also pointed out the importance of this issue. As an example, \cite{kantarciouglu2011classifier} use a Stackelberg game in which both players know each other payoff functions. Only \cite{grosshans2013bayesian} have attempted to relax common knowledge assumptions in adversarial regression settings, reformulating the corresponding problem as a Bayesian game.

In this paper we present a novel framework for AC based on Adversarial Risk Analysis (ARA), \cite{adversarialRiskAnalysis2009}. This is an emergent paradigm supporting decision makers who confront adversaries in problems with random consequences that depend on the actions of all participants. ARA provides one-sided prescriptive support to a decision maker maximizing her subjective expected utility by treating the adversaries' decisions as random variables. To forecast them, we model the adversaries' problems; 
however, our uncertainty about their probabilities and utilities is propagated leading to the corresponding random optimal adversarial decisions which provide the required distributions. ARA operationalizes the Bayesian approach to games, \cite{kadane1982subjective} and \cite{raiffa1982art}, facilitating a procedure to predict adversarial decisions. Compared with standard game theoretic approaches, ARA does not assume the standard common knowledge hypothesis, according to which agents share information about utilities and probabilities. Thus, we propose ACRA, an approach to robustify classification in adversarial settings based on ARA, which stems from the pioneering work by \cite{adversarialClassification2004} but avoids common knowledge assumptions prevalent in the available literature.


\section{Adversarial Classification based on Adversarial Risk Analysis}
\label{sec:acra}

In binary classification settings, an agent, that we call classifier ($C$, she), may receive two types of objects, denoted as malicious ($y=+$) or innocent ($y=-$). Objects have features $x$ whose distribution depends on their type $y$. Classification problems can be broken down into two separate stages, \cite{bishop}: an inference stage for learning $p_C(y|x)$, the classifier beliefs about the instance type given the features; and a decision stage in which the agent, based on these posterior probabilities, makes a class assignment decision $y_C$ perceiving some utility $u_C(y_C,y)$. The agent decides by maximizing expected utility, \cite{french2000statistical}. This problem may be formulated through an influence diagram (ID), \cite{jensen2012information}, as in Figure \ref{fig:classification}. Square nodes describe decisions; circle nodes, uncertainties; double nodes represent deterministic aspects; and finally, hexagonal nodes refer to the associated utilities. Arcs have the same interpretation as in \cite{evaluatingInfluenceDiagrams1986}; those arcs pointing to decision nodes are dashed and represent information available when the corresponding decisions are made.
\pagebreak

\begin{center}
\begin{figure}[h!]
\subfloat[Classification as an Influence Diagram \label{fig:classification}]
{
\begin{tikzpicture}[->, >=stealth', shorten >=    1pt, auto, node distance = 3cm, semithick, scale = 0.75, transform shape]
  \node[decisionC ](C) []                         {\footnotesize $y_C$};
  \node[chanceC](Y) [above left of = C]            {\footnotesize $y$};
  \node[chanceC](X) [right of = Y]           {\footnotesize $x$};
  \node[utilityC ](UC)[below left of = C ]        {\footnotesize $u_C$};
  \path
  (Y) edge[out =   0, in =  180        ] node {}(X)
  (Y) edge[out =   -90, in = 90        ] node {}(UC)
  (X) edge[dashed] node {}(C)
  (C) edge[out =   -135, in =  45, ] node {}(UC);
  \end{tikzpicture}
}\hfill
\subfloat[Adversarial classification as a Bi-agent Influence Diagram \label{fig:jointProblem}]
{
   \begin{tikzpicture}[->, >=stealth', shorten >= 1pt, auto, node distance = 3cm, semithick, scale = 0.75, transform shape]
 \node[decisionC ](C) []                         {\footnotesize $y_C$};
 \node[chance,accepting](XS) [right of = C]      {\footnotesize $x'$};
 \node[chance](Y) [above left of = C]            {\footnotesize $y$};
 \node[chance](X) [above left of = XS]           {\footnotesize $x$};
 \node[decisionA](A) [above right of = XS]       {\footnotesize $a$};
 \node[utilityC ](UC)[below left of = C ]        {\footnotesize $u_C$};
 \node[utilityA ](UA)[below right of = XS ]      {\footnotesize $u_A$};

 \path
 (Y) edge[out =   0, in =  180        ] node {}(X)
 (Y) edge[out =   -90, in = 90        ] node {}(UC)
 (Y) edge[out =   -35, in =  145] node {}(UA)
 (Y) edge[out = 45, in = 135, dashed] node {}(A)
 (X) edge[out =   -45, in = 135, ] node {}(XS)
 (X) edge[out =   0, in =  -180, dashed] node {}(A)
 (A) edge[out =  -135 , in = 45, ] node {}(XS)
 (A) edge[out =   -90, in =  90, ] node {}(UA)
 (C) edge[out =   -135, in =  45, ] node {}(UC)
 (C) edge[out =   -20, in =  160, ] node {}(UA)
 (XS) edge[out =   180, in =  0, dashed] node {}(C)
 ;
 \end{tikzpicture}
}
\caption{}
\end{figure}
\end{center}

In adversarial settings, another agent, called adversary ($A$, he), chooses an attack $a$ which, applied to the features $x$, leads to the perturbed data $x'=a(x)$ that are actually observed by $C$. A general transformation from $x$ to $x'$ will be designated $a_{x \rightarrow x'}$. In this case, the ID describing the classification problem must be augmented to incorporate adversarial decisions, leading to the bi-agent influence diagram (BAID), \cite{koller2003multi}, in Figure \ref{fig:jointProblem}. Grey nodes refer to issues solely affecting $A$'s decision; white nodes to issues solely pertaining to $C$'s decision; finally, striped nodes affect both agents. The adversary and classifier decisions are represented through nodes $a$ (chosen attack) and $y_C$ (classification choice), respectively. The impact of the data transformation over $x$ implemented by $A$ is described through node $x'$. The utilities of $A$ and $C$ are represented with nodes $u_A$ and $u_C$, respectively. Upon observing a particular $x'$, $C$ needs to determine the object class $y$. Her guess $y_C$, which we shall also denote $c(x')$, provides her with utility $u_C (y_C, y)$. As before, she aims at maximizing expected utility. However, $A$ also aims at maximizing his expected utility trying to confuse the classifier. His utility has the form  $u_A(y_C,y,a)$, when $C$ says $y_C$, the actual label is $y$ and the attack is $a$, which has an implementation cost.

\cite{koller2003multi} illustrated how to compute Nash equilibria in generic BAIDs under common knowledge conditions. \cite{GONZALEZORTEGA20191085} extended BAIDs to situations in which there is no common knowledge, by using the ARA solution concept, \cite{banks2015adversarial}, providing algorithms to compute ARA solutions in BAIDs and displaying generic interactions between two agents over time in support of one of the agents, the Defender, when protecting from the other, the Attacker. The graph-theoretic ideas which are core in that paper, are not required here as we have just one decision point for each of the agents, in a sequential fashion. We are actually interested in a problem, adversarial classification, different from the ones presented in such paper, and we provide a thorough study of how ARA can be used to address it. As a consequence, we adopt new forms and models adapted to classification tasks as well as specific algorithms to cope with the large size problems considered, paying special attention to robustness issues.


In this paper, we thus develop a framework to support the classifier $C$ in choosing her classification decision and studying its robustness to exploratory attacks, defined to have influence just over the operational data, but not over the training data. As in \cite{adversarialClassification2004}, we study cases in which $A$ does not attack innocent instances ($y = -$), denominated \textit{integrity-violation attacks}, as is the case in most security scenarios. We restrict our attention to deterministic attacks, in the sense that their output is not random. \cite{AdversarialMachineLearning2011} and \cite{Barreno2006} provide taxonomies of attacks against classifiers. 

We shall need to forecast the attacker's actions to support $C$ in her decision making process. For that we consider his problem. As we lack common knowledge, we shall model our uncertainty about $A$'s beliefs and preferences and compute $A$'s random optimal attack, which provides the required forecasting distribution.

\subsection{The classifier problem} \label{subsec:theClassifierProblem}
We present first the classification problem faced by $C$ as a Bayesian game in Figure \ref{fig:classifierProblem}, deduced from Figure \ref{fig:jointProblem}. We formulate a decision problem for $C$ in which $A$'s decision appears as random to the classifier, since she does not know how the adversary will attack the data. Section \ref{subsec:theAdversaryProblem} provides a procedure to estimate the corresponding probabilities making this approach operational.

Suppose for now that we are capable of assessing from the classifier:
\begin{center}
\begin{figure}[h]
\subfloat[Classifier problem \label{fig:classifierProblem}]
{
\begin{tikzpicture}[->, >=stealth', shorten >= 1pt, auto, node distance = 3cm, semithick, scale = 0.75, transform shape]
 \node[decisionC ](C) []                         {\footnotesize $y_C$};
 \node[chance,accepting](XS) [right of = C]      {\footnotesize $x'$};
 \node[chance](Y) [above left of = C]            {\footnotesize $y$};
 \node[chance](X) [above left of = XS]           {\footnotesize $x$};
 \node[chanceA](A) [above right of = XS]       {\footnotesize $a$};
 \node[utilityC ](UC)[below left of = C ]        {\footnotesize $u_C$};

 \path
 (Y) edge[out =   0, in =  180        ] node {}(X)
 (Y) edge[out =   -90, in = 90        ] node {}(UC)
 (Y) edge[out = 45, in = 135] node {}(A)
 (X) edge[out =   -45, in = 135, ] node {}(XS)
 (X) edge[out =   0, in =  -180, ] node {}(A)
 (A) edge[out =  -135 , in = 45, ] node {}(XS)
 (C) edge[out =   -135, in =  45, ] node {}(UC)
 (XS) edge[out =   180, in =  0, dashed] node {}(C)
 ;
 \end{tikzpicture}
}\hfill
\subfloat[Adversary problem \label{fig:adversaryProblem}]
{
 \begin{tikzpicture}[->, >=stealth', shorten >= 1pt, auto, node distance = 3cm, semithick, scale = 0.75, transform shape]
 \node[chanceC ](C) []                         {\footnotesize $y_C$};
 \node[chance,accepting](XS) [right of = C]      {\footnotesize $x'$};
 \node[chance](Y) [above left of = C]            {\footnotesize $y$};
 \node[chance](X) [above left of = XS]           {\footnotesize $x$};
 \node[decisionA](A) [above right of = XS]       {\footnotesize $a$};
 \node[utilityA ](UA)[below right of = XS ]      {\footnotesize $u_A$};

 \path
 (Y) edge[out =   0, in =  180        ] node {}(X)
 (Y) edge[out =   -35, in =  145        ] node {}(UA)
 (Y) edge[out = 45, in = 135, dashed] node {}(A)
 (X) edge[out =   -45, in = 135, ] node {}(XS)
 (X) edge[out =   0, in =  -180, dashed] node {}(A)
 (A) edge[out =  -135 , in = 45, ] node {}(XS)
 (A) edge[out =   -90, in =  90, ] node {}(UA)
 (C) edge[out =   -20, in =  160, ] node {}(UA)
 (XS) edge[out =   180, in =  0, ] node {}(C)
 ;
 \end{tikzpicture}
}
\caption{}\label{id}
\end{figure}
\end{center}
\begin{enumerate}
\item $p_C(y)$, which describes her beliefs about the class distribution, with $p_C(+)\,+ \,p_C(-)=1$, and $p_C(+),p_C(-) \geq 0$.
\item $p_C(x|y)$, modeling her beliefs about the feature distribution given the class, when $A$ is not present. Thus, we need $p_C(x|+)$  and $p_C(x|-)$. Since we focus on exploratory attacks, we can estimate $p_C(x|y)$ and $p_C(y)$ training a generative classifier, \cite{bernardo2007generative}, on data which is clean by assumption. 
\item $p_C(x'|a,x)$, which models her beliefs about the transformation results. Since we consider only deterministic transformations, it will actually be the case that $p_C(x'|a,x)= I( x' = a(x) )$, where $I$ is the indicator function.
\item $u_C(y_C,y)$, describing $C$'s utility when she classifies as $y_C$ an instance whose actual label is $y$. 
\item $p_C(a|x, y)$, portraying $C$'s beliefs about $A$'s action, given $x$ and $y$.
\end{enumerate}
In addition, we assume that $C$ is able to compute the set $\mathcal{A}(x)$ of possible attacks over a given instance $x$. When she observes $x'$, she could compute the set $\mathcal{X}' = \lbrace x : a(x) = x' ~\text{for}~\text{some}~a \in \mathcal{A}(x)\rbrace$ of instances potentially leading to $x'$. She should then aim at choosing  the class $y_C$ with maximum posterior expected utility. In our context, this means that she must find the class $c(x')$ such that 
\begin{eqnarray*}
c(x')&=&\argmax_{y_C} \sum_{y \in \lbrace +,- \rbrace} u_C(y_C, y) p_C(y|x')=
  \\
&=& \argmax_{y_C} \sum_{y \in \lbrace +,- \rbrace} u_C(y_C, y) p_C(y) p_C(x'|y) =\\
&=& \argmax_{y_C} \sum_{y \in \lbrace +,- \rbrace}  u_C(y_C, y) p_C(y)  \sum_{x \in \mathcal{X}'} \sum_{a \in \mathcal{A}(x)} p_C(x', x, a| y).
\end{eqnarray*}
For the second equation, we apply Bayes formula to compute $p(y|x')$, but ignore the denominator, which is irrelevant for optimization purposes. Then, we expand the term $p(x'|y)$ taking into account the possible attacks. The presence of $A$ thus modifies $p_C(x'|y)$, preventing us from directly using the training set estimates of these elements. Therefore, we need to take into account $A$'s modifications through the probabilities $p_C(x', x, a| y)$. Furthermore, expanding the last expression, we have \small
\begin{eqnarray*}
&c(x')& = \argmax_{y_C} \sum_{y \in \lbrace +,- \rbrace} \bigg[ u_C(y_C, y) p_C(y) \sum_{x \in \mathcal{X}'} \sum_{a \in \mathcal{A}(x)} p_C(x'| x,a,y) p_C(x,a | y)\bigg] = \\
&=& \argmax_{y_C} \sum_{y \in \lbrace +,- \rbrace} \bigg[ u_C(y_C, y) p_C(y) \sum_{x \in \mathcal{X}'} \sum_{a \in \mathcal{A}(x)} p_C(x'| x,a) p_C(a |x,y) p_C(x|y)\bigg] .
\end{eqnarray*} \normalsize
Recalling now that we consider only integrity-violation attacks, we have $p_C(a|x, -)$ $= I(a = \textit{id})$, where \textit{id} stands for the identity attack leaving $x$ unchanged and $I$ is the indicator function. Then, taking also into account assumption 3, the problem to be solved by $C$ is \small
\begin{eqnarray}
 c(x') &=& \argmax_{y_C} \bigg[ u_C(y_C, +) p_C(+) \sum_{x \in \mathcal{X}'} \sum_{a \in \mathcal{A}(x)} I(x' = a(x) ) p_C(a |x,+) p_C(x|+)  \nonumber\\
  &+& u_C(y_C, -) p_C(-) \sum_{x \in \mathcal{X}'} \sum_{a \in \mathcal{A}(x)} I(x' = a(x) ) I(a = \textit{id})p_C(x|-)\bigg] = \nonumber\\
  &=& \argmax_{y_C} \bigg[ u_C(y_C, +) p_C(+) \sum_{x \in \mathcal{X}'} p_C(a_{x \rightarrow x'} |x,+) p_C(x|+) \nonumber \\
  &+& u_C(y_C, -) p_C(x'|-)p_C(-)\bigg], \label{pis}
\end{eqnarray} \normalsize
where $p_C(a_{x \rightarrow x'} |x,+)$ designates the probability that $A$ will execute an attack that transforms $x$ into $x'$, when he receives $(x,y=+)$, according to $C$. 

Note that if the above mentioned game-theoretic common knowledge assumptions held, $C$ would be able to compute the set of instances that, with probability 1, $A$ would change to $x'$: we would know $A$'s beliefs and preferences and, therefore, would be able to compute the attacks he is actually implementing. Thus, we find out that for those potential attacks, $p_C(a_{x \rightarrow x'} |x,+)$ would be 1, and 0 for every other. Then, the model would just take into account instances with probability 1, ignoring the others. But common knowledge is not available, so we actually lack $A$'s beliefs and preferences. ACRA models our uncertainty around them. As we shall see, in doing so we enhance model robustness. Notice that in \eqref{pis}, we are summing $p_C(x|+)$ over all possible originating instances, weighting each element by $p_C(a_{x \rightarrow x'} |x,+)$, thus taking into account the uncertainty about $A$'s decision problem.


The ingredients 1-4 required in the analysis are standard in the decision analytic practice, \cite{clemen2013making}. However, the fifth element $p_C(a_{x \rightarrow x'}|x, y)$, demands strategic thinking from $C$. To facilitate the corresponding forecast and make the approach operational, we consider $A$'s decision making process.

\subsection{The attacker problem}
\label{subsec:theAdversaryProblem}
We assume that $A$ aims at modifying $x$ to maximize his expected utility by making $C$ classify malicious instances as innocent. The decision problem faced by $A$ is presented in Figure \ref{fig:adversaryProblem}, deduced from Figure \ref{fig:jointProblem}. In it, $C$'s decision appears as an uncertainty to $A$. Suppose for now that we have available from him:
\renewcommand{\labelenumi}{\arabic{enumi}'.}{
\begin{enumerate}
\item $p_A(x'|a,x)$, describing his beliefs about the transformation results. As with $C$, we make $p_A(x'|a,x)$ $ = I(x' = a(x))$.
\item $u_A(y_C ,y  , a)$, which describes the utility that $A$ attains when $C$ says $y_C$, the actual label is $y$ and the attack is $a$. Note that this reflects certain attack implementation costs.
\item $p_A(c(x')|x')$, which models $A$'s beliefs about the classification result when $C$ observes $x'$. 
\end{enumerate}
}
\noindent In connection with the last ingredient, let us designate by $p=p_A(c (a (x)) = + | a (x) )$ the probability that $A$ concedes to $C$ saying that the instance is malicious, given that she observes the attacked data $x' = a(x)$. Since he will have uncertainty about it, we denote its density by $f _A (p | a(x) )$ with expectation $p_{a(x)} ^A $. 

Among all attacks, $A$ would choose that maximizing his expected utility
\begin{eqnarray}
a^*(x, y) &=& \argmax_{a} \int \bigg[ u_A(c(a(x)) = +,y,a) \cdot p \nonumber\\
&+& u_A(c(a(x)) = -,y,a) \cdot (1-p)\, \bigg] f _A (p | a(x) )dp.
\label{attackerprob}
\end{eqnarray}
Since we assume that $A$ does not change the data when $y=-$, we only consider the case $y=+$. Then, $A$'s expected utility when he adopts attack $a$ and the instance is $(x,y=+)$ will be
\begin{eqnarray*}
\int \bigg[ u_A( +, +, a) \, \, p  +
       u_A( -, +, a) \, \, (1 - p) \bigg] f_A (p | a(x) ) dp = \\
= \left[u_A( +, +, a) - u_A( -, +, a)\right] p_{a(x)}^A  +  u_A( -, +, a) .
\end{eqnarray*}

However, the classifier does not know the involved utilities $u_A$ and probabilities $ p_{a(x)} ^A $ from the adversary. Suppose we may model her uncertainty through a random utility function $U _A$ and a random expectation $ P_{a(x)}^A $. Then, we could solve for the random optimal attack, optimizing the random expected utility
\begin{eqnarray*}
 A^*(x, +)=\argmax_{a} \bigg(  \left[ U_A( +, +, a) - U_A( -, +, a) \right] P_{a(x)}^A   +    U_A( -, +, a) \bigg),
\end{eqnarray*}
and make $p_C(a_{x \rightarrow x'}|x,+) =Pr(A^*(x,+)=a_{x \rightarrow x'})$, assuming that the set of attacks is discrete. 

We have therefore provided a way to approximate the remaining fifth ingredient in the classifier problem in Section \ref{subsec:theClassifierProblem}, which is now operational. In general, to approximate it we use simulation drawing $K$ samples $\bigl(U_A^k(y_C , +, a), P_{a(x)} ^{A,k} \bigr)$, $k = 1,\dots,K$ from the random utilities and probabilities, finding
\begin{eqnarray*}
A^*_k (x, +)=\argmax_{a}
\bigg( \left[   U_A^k( +, +, a) - U_A^k( -, +, a) \right] P_{a(x)}^{A,k}   +    U_A^k( -, +, a) \bigg)
\end{eqnarray*}
and estimating it through
\begin{eqnarray}
\widehat{p _C} ( a_{x \rightarrow x'}\,|\,x, +) =  \frac {\# \{A_k^*(x, +) =  a_{x \rightarrow x'}\}} {K}. \label{mc_estimate}
\end{eqnarray}
 It is easy to prove that \eqref{mc_estimate} converges almost surely to $p_C(a_{x \rightarrow x'}|x,+)$.

Of the required random elements, it is relatively easy to model the random utility $U_A (y_C , +, a)$ which would typically include two components. The first one refers to $A$'s gain from $C$'s decision. If we adopt the notation $Y_{y_C y}$ to represent the gain when $C$ decides $y_C$ and the actual label is $y$, we use: $-Y_{++} \sim Ga (\alpha _1, \beta _1)$ with $\alpha _1/\beta _1=-d$, $d$ being the expected gain for $A$ and variance $\alpha _1 / \beta _1 ^2$ as perceived, thus assuming that the utility obtained by $A$ when $C$ classifies a truly malicious instance as malicious is negative. Similarly, $Y_{-+} \sim Ga (\alpha _2, \beta _2)$ with $\alpha _2/\beta _2=e$, the expected gain for $A$, and variance $\alpha _2 / \beta _2 ^2$ as perceived. Our choice of gamma distributions to model the uncertainty about the adversary preferences is motivated by its properties: they combine the relative simplicity of depending on two parameters with the variety of shapes they can take and the possibility of easily specifying their parameters once a guess (e.g., the mean) about the quantity of interest is available along with an opinion about the spread (e.g., the variance). Finally, $Y_{+-}= Y_{--}= \delta_0$, the degenerate distribution at 0, assuming that the adversary does not receive any utility from innocent instances. The second component refers to the random cost $B$ of implementing an attack. Then, the gain of the attacker would be $Y_{y_C y}-B$. Finally, assuming that $A$ is risk prone, \cite{french2000statistical}, the random utility could be computed as $U_A (y_C , y, a)= \exp ( \rho \, (Y _{y_C y}-B))$ with, say, $\rho \sim U [a_1,a_2]$, $a_1 > 0$, the random risk proneness coefficient.


On the other hand, modeling $P^A _{a(x)}$, $A$'s (random) expected probability that $C$ declares an instance as malicious when she observes $x'=a(x)$, is more delicate. It entails strategic thinking as $C$ needs to understand his opponent's beliefs about what classification she will make when she observes $x'$. This could be the beginning of a hierarchy of decision making problems, as described in \cite{araCounter} in a much simpler context. We illustrate here the initial stage of such hierarchy in our problem area. First, $A$ does not know the terms in the decision making problem \eqref{pis} faced by the classifier. By assuming uncertainty over them through the random distributions $P_C^A (+)$, $P_C ^A (x|+)$, $P_C ^A (x'|-)$, $P_C^A (a_{x \rightarrow x'}|x,+)$ and utilities $U _C ^A (y_C,+)$, $U_C ^A (y_C,-)$, he would get the corresponding random optimal decision replacing the incumbent elements in \eqref{pis} to obtain $P^A _{a(x)}$. However,
observe that this requires the assessment of $P_C^A (a_{x \rightarrow x'}|x,+)$ (what $C$ believes that $A$ thinks about her beliefs concerning the action he would implement given the observed data) for which there is a strategic component,
leading to the next stage in the announced hierarchy. One would typically stop at a level in which no more information is available. At that stage, we could use a non-informative prior over the involved probabilities and utilities.

For the first stage of this hierarchy, a relevant heuristic to assess $P^A _{a(x)}$ may be based on the probability $Pr _C (c (x') = +|x')=r$ that $C$ assigns to the object received being malicious assuming that she observed $x'$, with some uncertainty around it. 
Being a probability, $r$ ranges in $[0,1]$ and 
we could make $P_{a(x)}^A \sim \beta e (\delta _1, \delta _2 )$,with mean $\delta _1 / (\delta _1 + \delta _ 2) = r$ and variance $ (\delta _1 \delta _2) / [(\delta _1 + \delta _2 )^2 (\delta _1 + \delta _2 + 1) ]=var $ as perceived, leading to
 \begin{eqnarray}
 \delta_{1}=\left( \frac{1-r}{\textit{var}} - \frac{1}{r} \right) r^2, \hspace{1cm}
 \delta_{2}=\delta_{1}\bigg(\frac{1}{r}-1\bigg) . \label{deltas}
 \end{eqnarray}
Specifics would depend on the case at hand. In general, given the observed $x'$, we could consider all attacks leading to it, differentiating between instances with original label $+$ and those with original label $-$; computing the probabilities of observing the malicious ones and adding them to obtain $p_1$; performing the same process with innocent instances to obtain $p_2$; and, finally making $r=p_1 / (p_1+p_2)$.

\subsection{Algorithmic implementation} \label{subsec:Algorithm}

Once we have a Monte Carlo routine to estimate $p_C(a_{x \rightarrow x'} | x, +)$ as in \eqref{mc_estimate}, which entails the availability of a routine to generate from the random utility function, as in the Appendix, we implement the scheme described above as follows, where $\hat{x}$ indicates estimate of $x$.

\renewcommand{\labelenumi}{\arabic{enumi}.}
\begin{enumerate}
\item \textsc{Preprocessing}\\
Train a generative classifier to estimate $p_C(y)$ and $p_C(x|y)$, assuming that the training set has not been tainted.






\item \textsc{Operation} \\
Read $x'$.\\
\textsc{Estimate} $p_{C}(a_{x \rightarrow x'} |x,+)$.\\
Solve
\begin{eqnarray*}
c(x') &=& \argmax_{y_C} \bigg[ u(y_C, +) \widehat{p}_C(+) \sum_{x\in \mathcal{X}'} \widehat{p}_C(a_{x \rightarrow x'} |x,+) \widehat{p}_C(x|+) 
\\ &+& u(y_C, -) \widehat{p}_C(x'|-)\widehat{p}_C(-)\bigg].
\end{eqnarray*}

Output $c(x')$.

\end{enumerate}


\section{An example in spam detection}
\label{sec:conEx}
We illustrate the ACRA approach with a spam detection problem. We have data referring to $m$ emails characterized through the \textit{bag-of-words} representation,  \cite{zhang2010understanding}: binary features indicate the presence (1) or not (0) of $n$ relevant words in a dictionary. Additionally, a label indicates whether the message is spam $(+)$ or not $(-)$. Thus, an email is assimilated with an $n$-dimensional vector $x$ of 0's or 1's, together with a label $y$. In this example, we consider only good word insertion attacks, \cite{adversarialClassification2004}. For simplicity, we illustrate the method in detail only when adding at most one word (1-GWI), but our method can be applied, with precautions discussed later, to the case of $k$ added words ($k$-GWI). An example asseses how the method works in the case 2-GWI in Section \ref{sec:Application}. 1-GWI entails converting at most one of the 0's of the originally received message into a 1.


Given a message $x= (x_1, x_2,.., x_n)$, with $x_i \in \{ 0, 1 \}$, let us designate by $I(x)$ the set of indices such that $x_i=0$. Then, the set of possible attacks in this case is $\mathcal{A}(x) = \{ a_0=id, a_i; \forall i \in I(x) \}$, where $a_i$ transforms the $i$-th 0 into a 1. In turn, given a message $x'$ received by $C$, we designate by $J(x')$ the indices of features with value 1 in $x'$. If we designate by $x'_{j}$ a message potentially leading to $x'$, derived by changing the $j$-th 1 in $x'$ with a 0, the set of possibly originating messages would be $\mathcal{X}' = \lbrace x', x'_{j};  \forall j \in J(x') \rbrace$.

\subsection{Classifier elements}
The elements required to solve the classifier problem, Section \ref{subsec:theClassifierProblem}, include the utility function $u_C (y_C , y )$, which is standard, and the distributions $p_C(y)$ and $p_C(x|y)$, also standard if we just consider, as we do here, exploratory attacks. As we mentioned, we could use our favorite generative classifier to estimate them. Finally, $p_{C}(a_{x \rightarrow x'}|x,y)$ has a strategic component and we use ARA to approximate it.
\subsection{Adversary elements}

The adversary's random utilities follow the general arguments in Section \ref{subsec:theAdversaryProblem}. We also need to assess $P_{a(x)} ^A $. We use the heuristic there proposed, with the following caveat. If the original label is $ -$, the mail is innocent and the adversary does not change it, thus coinciding with the received one; we denote by $q _0 = p_C (x' | - ) p_C (-)$ the probability of this event happening, according to $C$. If the original label is $ +$, the mail is malicious, and $A$ might change it in an attempt to fool the classifier; according to $C$, the original message  $x'_{j}$ happens with probability $q_j= p_C (x'_{j} | + ) p_C (+), \forall j \in J(x')$. However, the adversary might decide not to attack even if the email is spam; according to $C$, this happens with probability $q_{n+1}=  p_C (x ' | + ) p_C (+ )$. Then
\begin{eqnarray}
r_a = \frac{\sum _{i \in J[a(x)]} q_i + q_{n+1}}{q_0 + \sum _{i \in J[a(x)]} q_i + q_{n+1}} \label{ra}
\end{eqnarray}
is the probability of $C$ believing that the observation $a(x)$ has label $+$, when she is aware of the presence of $A$. For such attack $a$, we could make $\delta _1^a / (\delta_1^a  + \delta _ 2^a ) =  r_a $ and $  (\delta _1^a \delta _2 ^a ) /  [(\delta _1 ^a + \delta _2^a  )^2 (\delta _1^a + \delta _2^a + 1)] = var$ and solve for $\delta_1^a$ and $\delta_2^a$ as in \eqref{deltas}.

\subsection{Example}
\label{sec:example}
The above ingredients allow us to implement Routines 1 and 2 in the Appendix, to generate from the random utility function and estimate $p_{C}(a_{x \rightarrow x'} |x,+)$, respectively. With this, we follow the scheme in Section \ref{subsec:Algorithm}.
%
%
%
We illustrate it in a simplified spam filtering problem\footnote{For the sake of reproducibility, we provide the open source version of the code used for the examples at \url{https://github.com/roinaveiro/ACRA_spam_experiment.git}. The data is publicly available at \url{https://archive.ics.uci.edu/ml/datasets/spambase}.
}, and compare it against the utility sensitive naive Bayes (NB) classifier, a standard non adversarial generative approach in this application area, \cite{song2009better}. We use the Spambase Data Set from the UCI Machine Learning repository, \cite{Lichman:2013}. It consists of 4601 emails, out of which 1813 are spam. For each email, the database contains information about 54 relevant words. The \textit{bag-of-words} representation with binary features assimilates each email with a 54 dimensional vector $x$ of 0's and 1's. The dataset will be divided into training and hold-out test sets, respectively comprising 75$\%$ and $25\%$ of the data. 

We first train a utility sensitive NB classifier using the training data, unaltered by assumption. For comparison purposes, and in order to check utility robustness, we use four utility functions. One of them is the 0/1 utility, i.e.\ the utility is 1 if the instance is correctly classified and 0 otherwise. For the rest, we chose utility 1 for correctly classified instances and -1 for spam classified as legitimate. The penalty for classifying non-spam mail as spam was set, respectively, to -2, -5 and -10 in the other three cases. The corresponding NB classifier will serve for comparison as well as basis for the ACRA approach providing the required $\widehat{p}_C(y)$ and $\widehat{p}_C(x|y)$.

To compare ACRA with NB on tampered data, we simulate attacks over the instances in the test set. For this purpose, we solved the adversary problem \eqref{attackerprob} for each test email. Uncertainty in the adversary's utility function is not present from his point of view, thus, we fixed $-u_A(+,+,a) = 5$, $u_A(-,+,a) = 5$, $u_A(-,-,a) = u_A(+,-,a) = 0$. The cost for implementing an attack was set to $b = 0.5 \cdot d(a)$, where $d(a)$ is the number of word changes (0 or 1) associated with attack $a$. The risk proneness coefficient was set to $\rho = 0.5$. Finally, the adversary would have uncertainty about $p_{a(x)}^A$, as this quantity depends on the classifier decision. We test ACRA against a worst case adversary, who knows the true value of $p_{a(x)}^A$. With this, we attacked each test email to generate the attacked test set. 

From the classifier's point of view, the adversary's parameters were fixed at: $-U_A(+,+,a) \sim Ga(\alpha_1, \beta_1)$ with $E[-U_A(+,+,a)] = 5$ and $\textit{Var}[-U_A(+,+,a)] = 0.01$, entailing $\alpha_1=  2500$, $\beta_1= 0.002$; $U_A(-,+,a) \sim Ga(\alpha_2, \beta_2)$, again with $\alpha_2= 2500 $, $\beta_2= 0.002 $; $U_A(-,-,a) = U_A(+,-,a) = \delta_0$. The random cost of implementing a particular attack $a$ was set to $B = d(a)\cdot \alpha$, where $d(a)$ is the number of word changes (0 or 1) associated with attack $a$, and $\alpha \sim U[0.4,0.6]$. The random risk proneness coefficient was set to $\rho \sim U[0.4,0.6]$. Observe that the attacker values are set as the means of the classifier distributions used to model them. We study later on how departures from the assumed adversary behaviour affects both ACRA and the corresponding game theoretic solution performance.

In order to obtain $P_{a(x)}^A$ for a given attack $a$, we need to generate from a beta distribution with mean $Pr _C (c (a(x)) = +|a(x))$, requiring its density to be concave in its support. Otherwise, we would be believing that the probability that $A$ concedes to $C$ deciding the instance $a(x)$ is malicious is peaked around 0 and 1 and low in between, which makes no sense in our context. Then, its variance must be bounded from above by $\Delta = \min \big \lbrace[r^2(1-r) ]/ (1+r), [r(1-r)^2] / (2-r) \big \rbrace$. We fix the adjustable variance $var$ at $k\Delta$ with $k \in [0,1]$: the bigger $k$ is, the bigger $C$'s uncertainty will be about $A$'s behavior. We ran experiments for each $k \in \lbrace 0.01, 0.1, 0.2, \cdots, 0.9 \rbrace$. Finally, we fixed $K=1000$, the Monte Carlo sample size in \eqref{mc_estimate}. 

\begin{figure}[h!]
    \centering
    \begin{minipage}{.5\textwidth}
        \centering
        \includegraphics[width=1\linewidth]{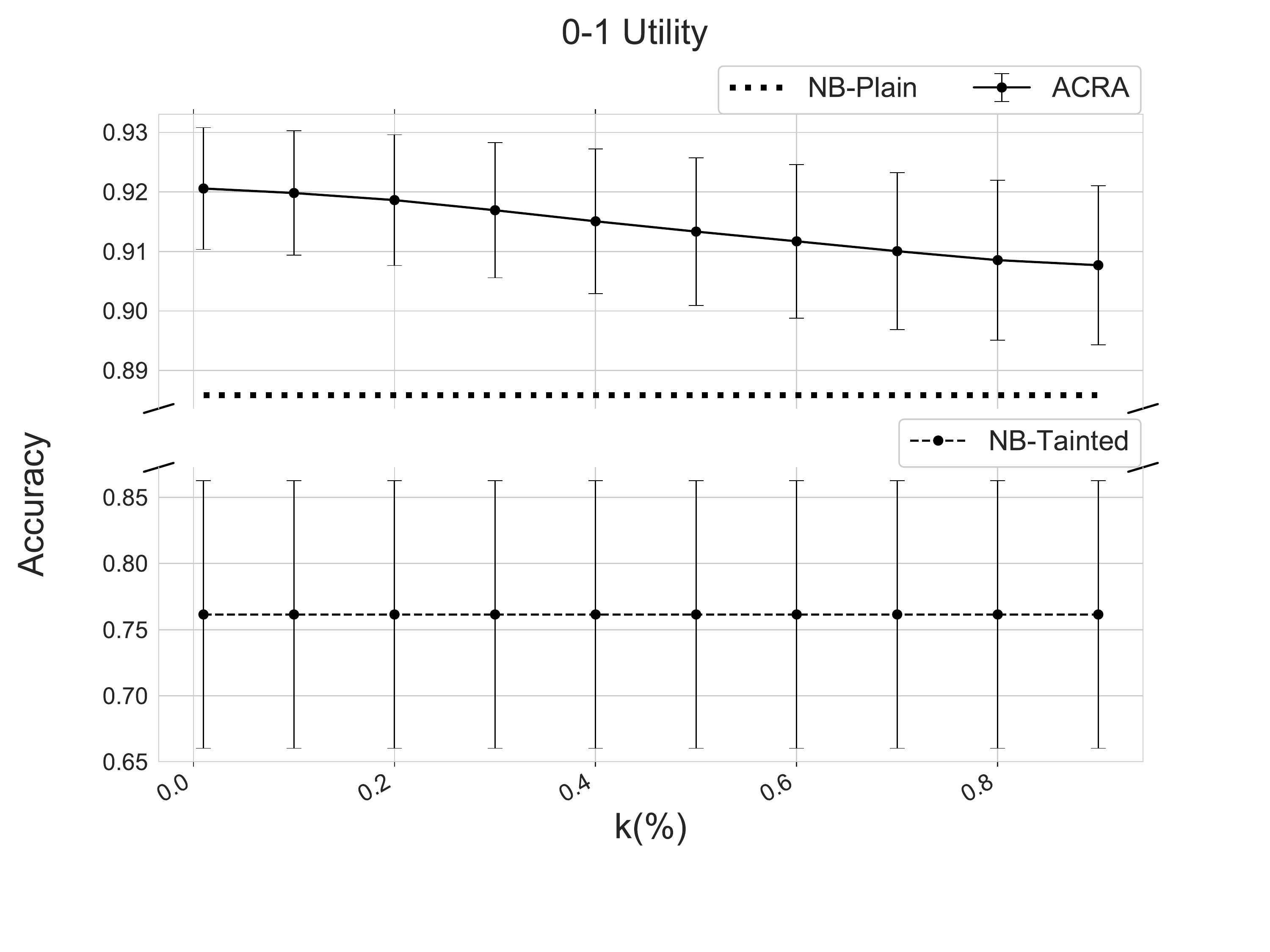}
    \end{minipage}%
    \begin{minipage}{0.5\textwidth}
        \centering
        \includegraphics[width=1\linewidth]{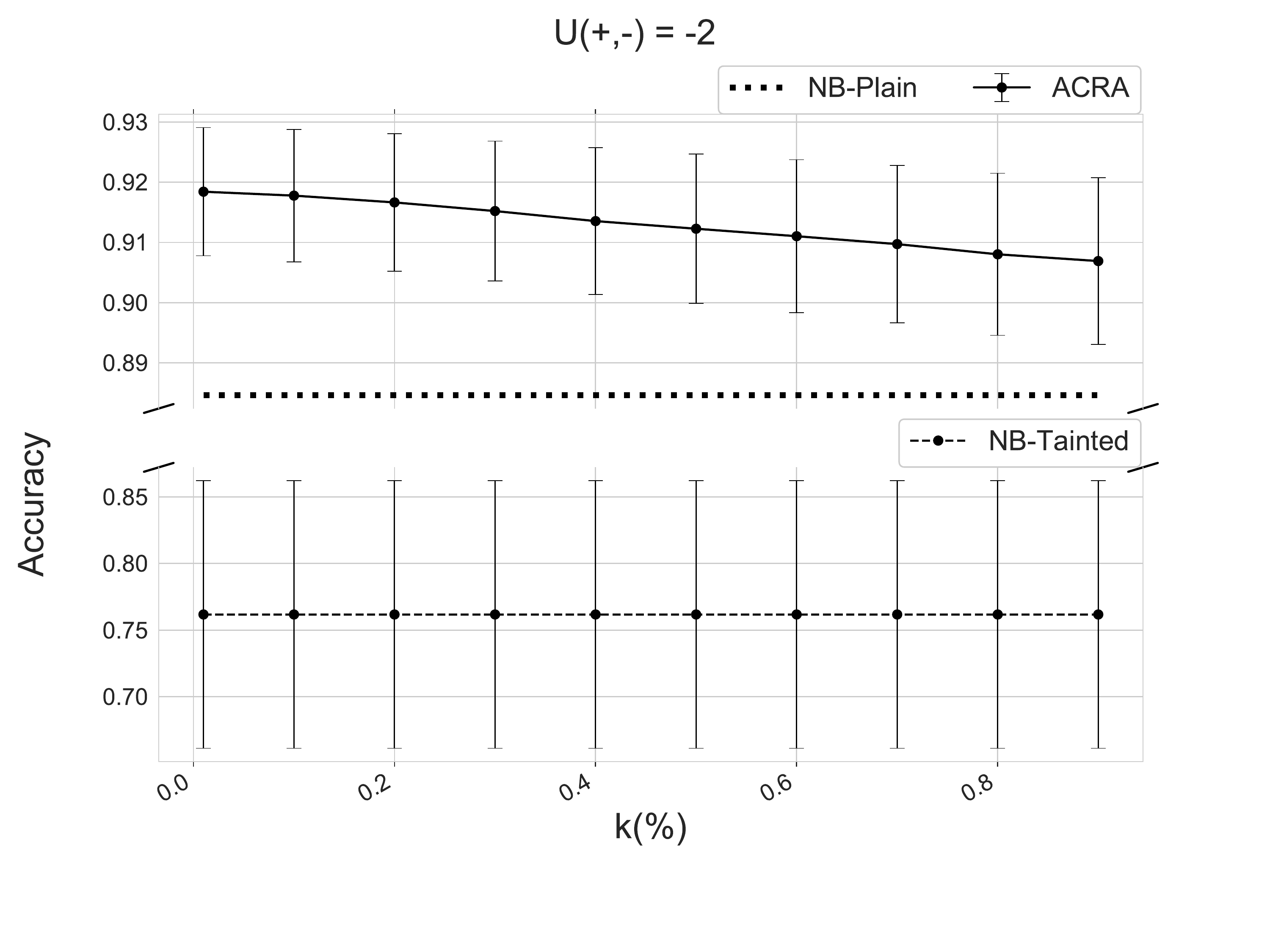}
    \end{minipage}
    \linebreak
    \begin{minipage}{.5\textwidth}
        \centering
        \includegraphics[width=1\linewidth]{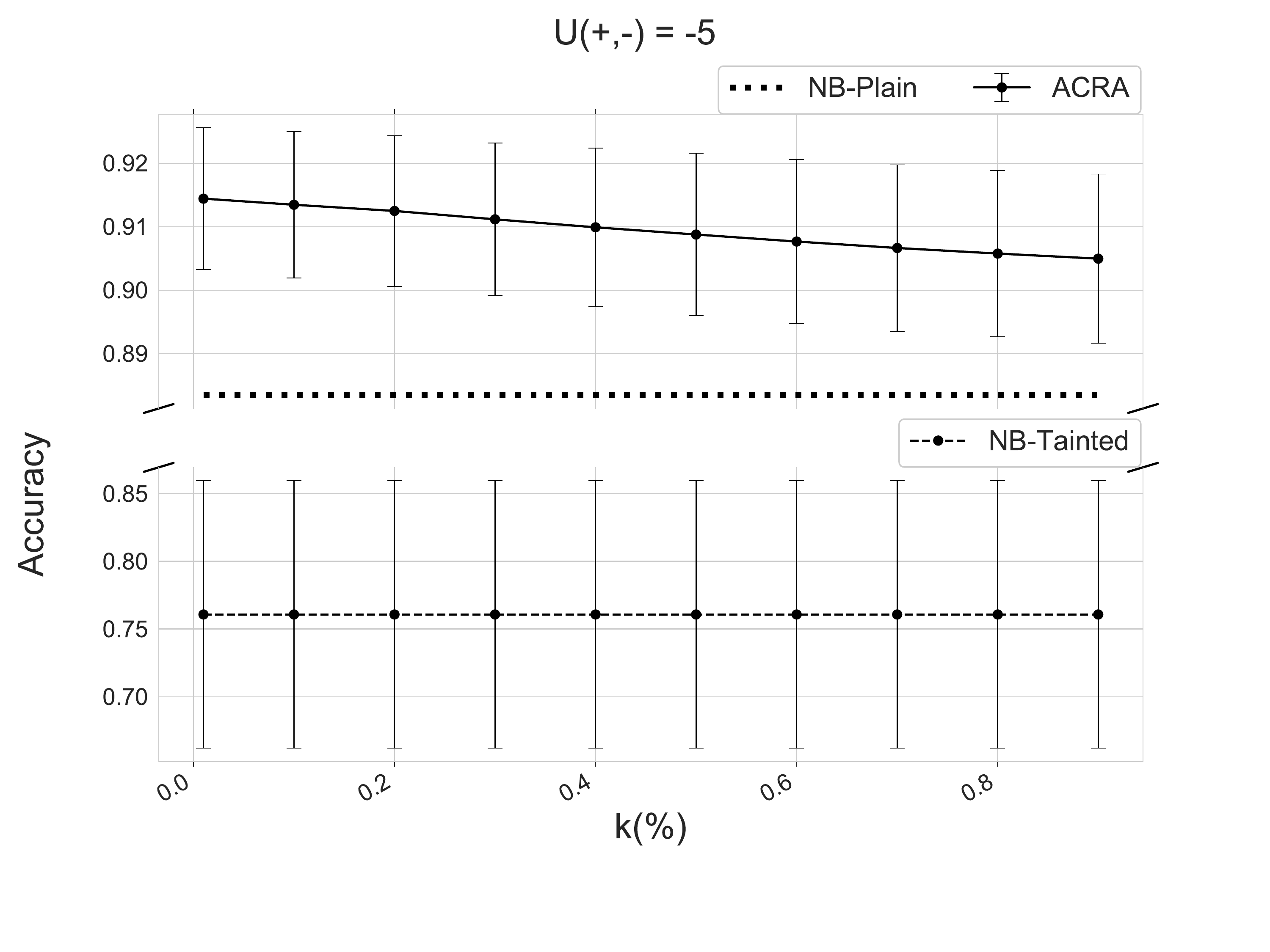}
    \end{minipage}%
    \begin{minipage}{0.5\textwidth}
        \centering
        \includegraphics[width=1\linewidth]{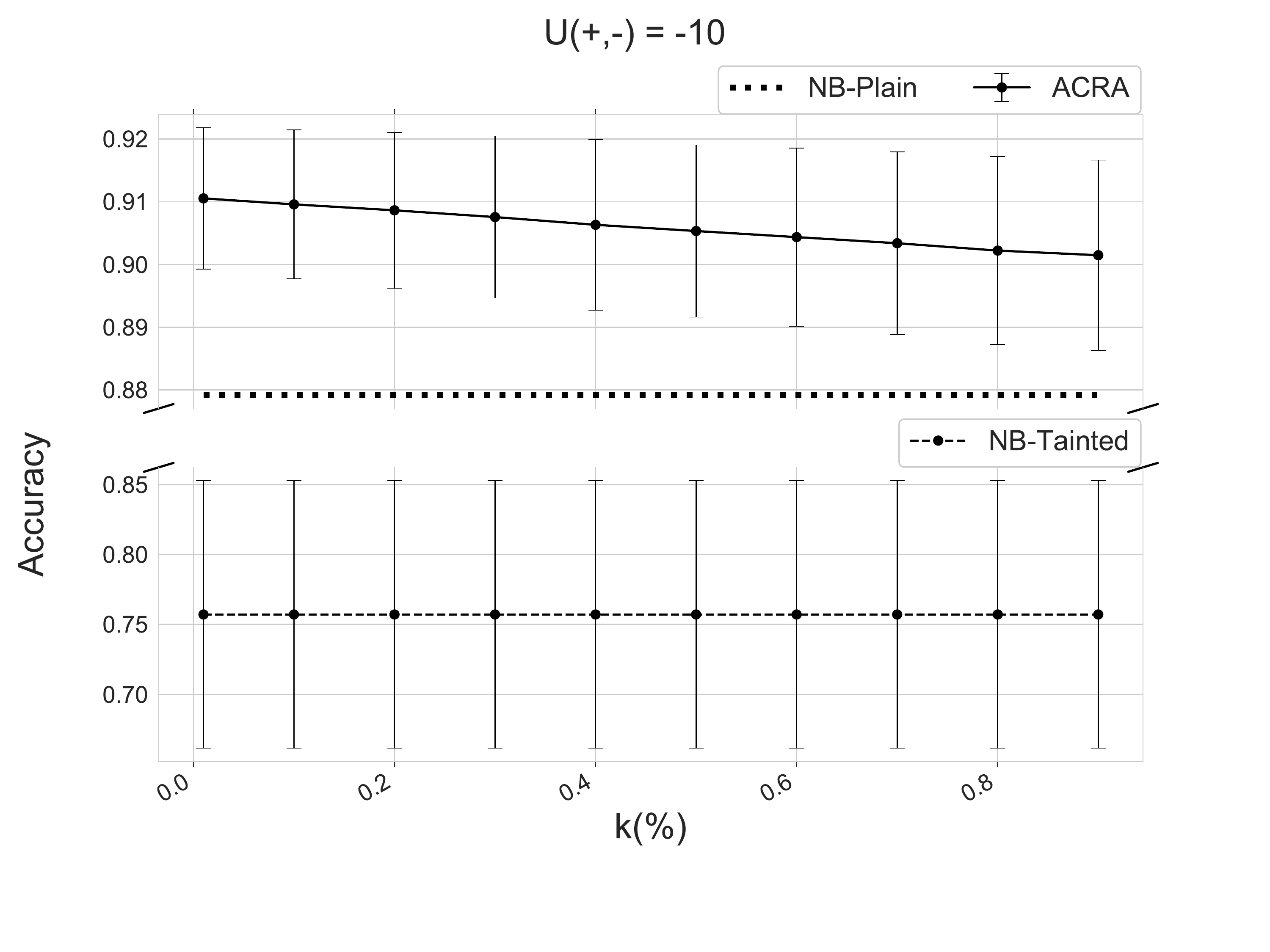}
    \end{minipage}
    \caption{Average accuracy versus $k$ for different utility models.}\label{acc}
\end{figure}

As performance metrics, we used the accuracy, utility, false positive (FPR) and false negative (FNR) rates, estimated via repeated hold-out validation over 100 repetitions, \cite{kim2009estimating}. We represent the results of the ACRA algorithm over the tampered test set with a solid line. The dashed line corresponds to the results of the utility sensitive NB on the attacked test, referred to as NB-Tainted. The error bars represent the standard deviation of each metric, also estimated through repeated hold-out validation. In addition, as a benchmark, we show the results of the utility sensitive NB over the original, untampered test set with a dotted line. We refer to these as NB-Plain. Obviously, NB-Plain and NB-Tainted metrics do not depend on $k$. 

\begin{figure}[h!]
    \centering
    \begin{minipage}{.5\textwidth}
        \centering
        \includegraphics[width=1\linewidth]{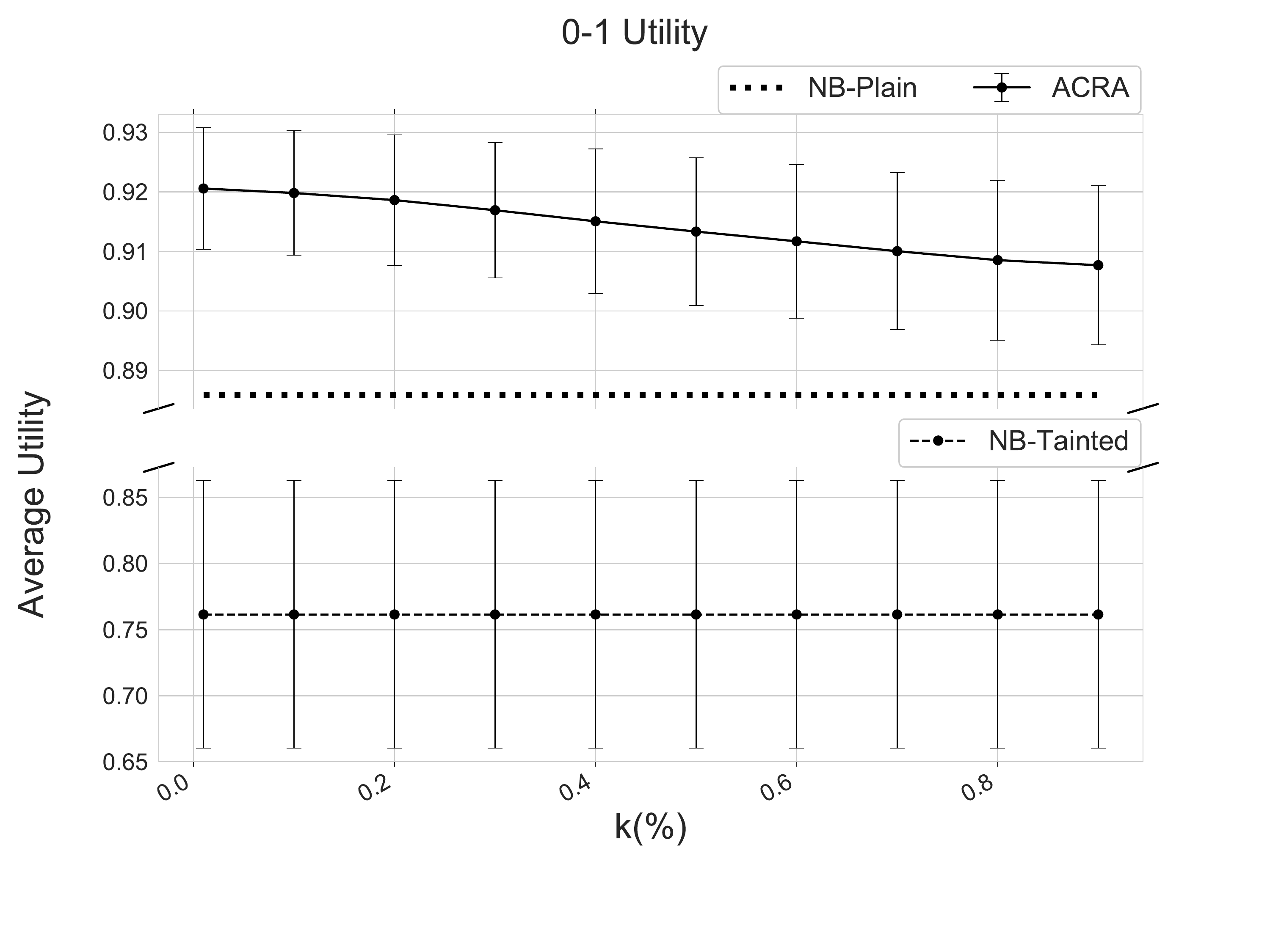}
    \end{minipage}%
    \begin{minipage}{0.5\textwidth}
        \centering
        \includegraphics[width=1\linewidth]{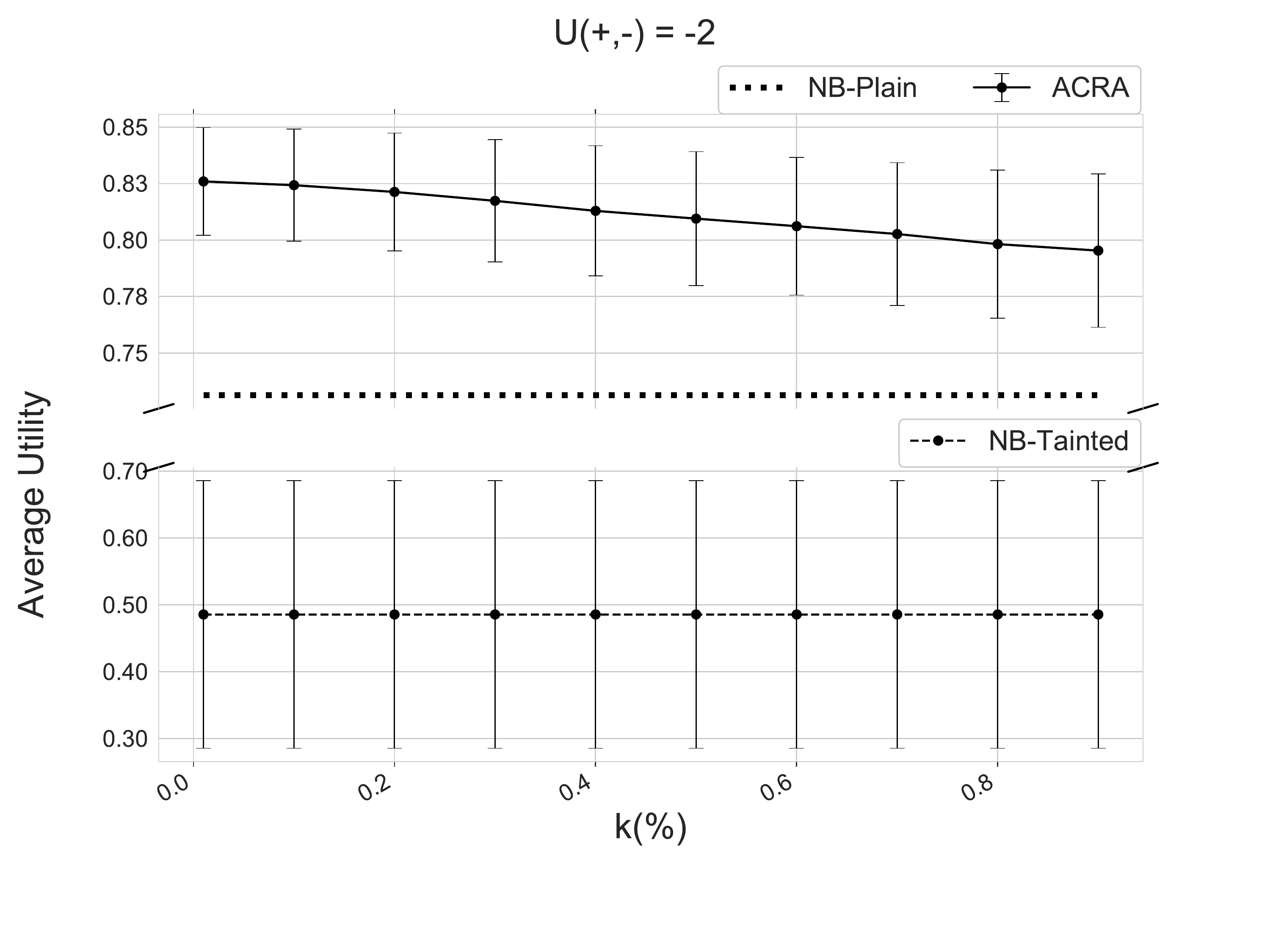}
    \end{minipage}
    \linebreak
    \begin{minipage}{.5\textwidth}
        \centering
        \includegraphics[width=1\linewidth]{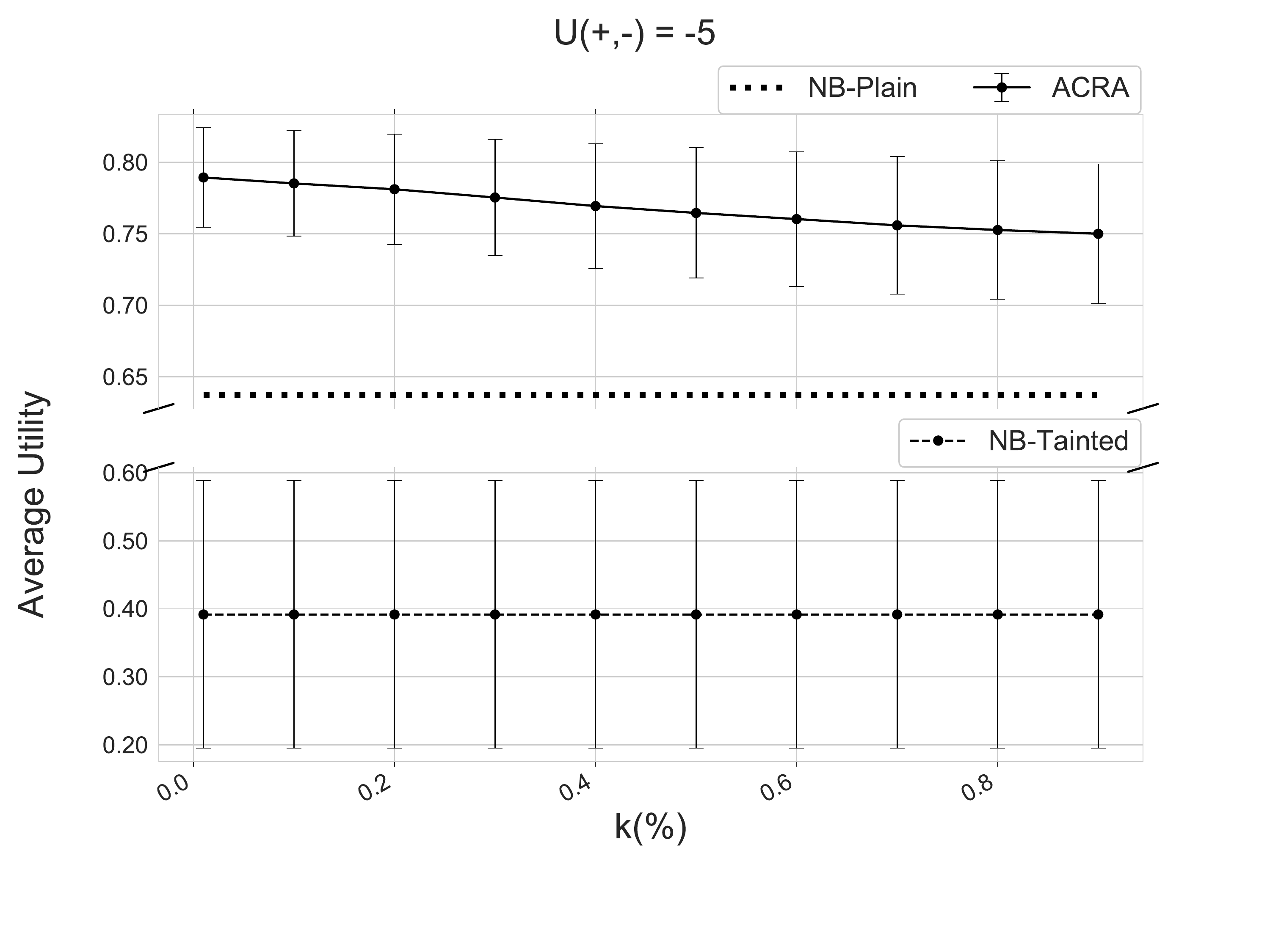}
    \end{minipage}%
    \begin{minipage}{0.5\textwidth}
        \centering
        \includegraphics[width=1\linewidth]{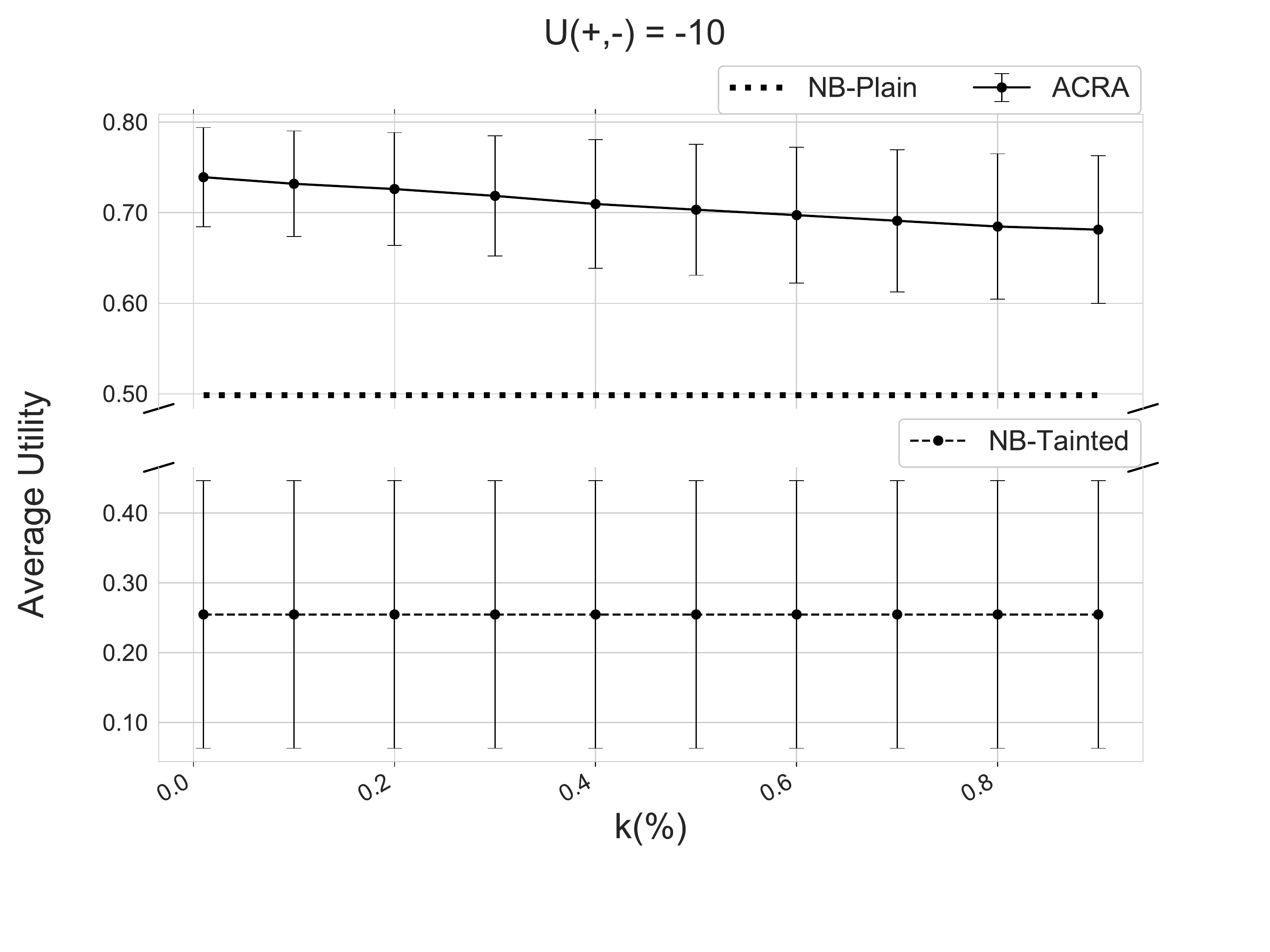}
    \end{minipage}
    \caption{Average attained utility versus $k$ for different utility models.}\label{avut}

\end{figure}

Figures \ref{acc} and \ref{avut} respectively present the average accuracies and utilities for various values of $k$ and the four utility models. Observe first that the presence of an adversary considerably degrades NB performance both in accuracy and average utility, as NB-Plain is consistently above NB-Tainted. This one is still correctly classifying the same proportion of non spam as NB-Plain, as such emails have not been attacked. However, NB-Tainted is not able to identify a large proportion of attacked spam emails. Consequently, as we increase the cost of misclassifying non-spam, reducing the relative importance of misclassifying spam, the performance of NB clearly degrades. This lack of robustness to attacks confirms the need to take into account the presence of adversaries.

In contrast, ACRA is robust to attacks and identifies most of the spam. Its overall accuracy is above $0.9$, thus identifying most non-spam emails. Observe though that ACRA degrades as $k$ grows: the bigger $k$ is, the less precise the knowledge that $C$ has about $A$ 
and the classifier performance will degrade. 
One of our contributions is providing parameters that may be tuned to adapt to the knowledge that the classifier could have about
her opponent. 

Very interestingly, note that in Figures \ref{acc} and \ref{avut} ACRA beats NB-Plain in both accuracy and utility. This effect has been observed by \cite{adversarialClassification2004} and \cite{goodfellow} for different algorithms and different application areas. The latter argues that taking into account the presence of an adversary has an effect similar to that of a regularizer, being able to improve the original accuracy of the base algorithm and making it more robust. 

\begin{figure}[ht]
    \centering
    \begin{minipage}{.5\textwidth}
        \centering
        \includegraphics[width=1\linewidth]{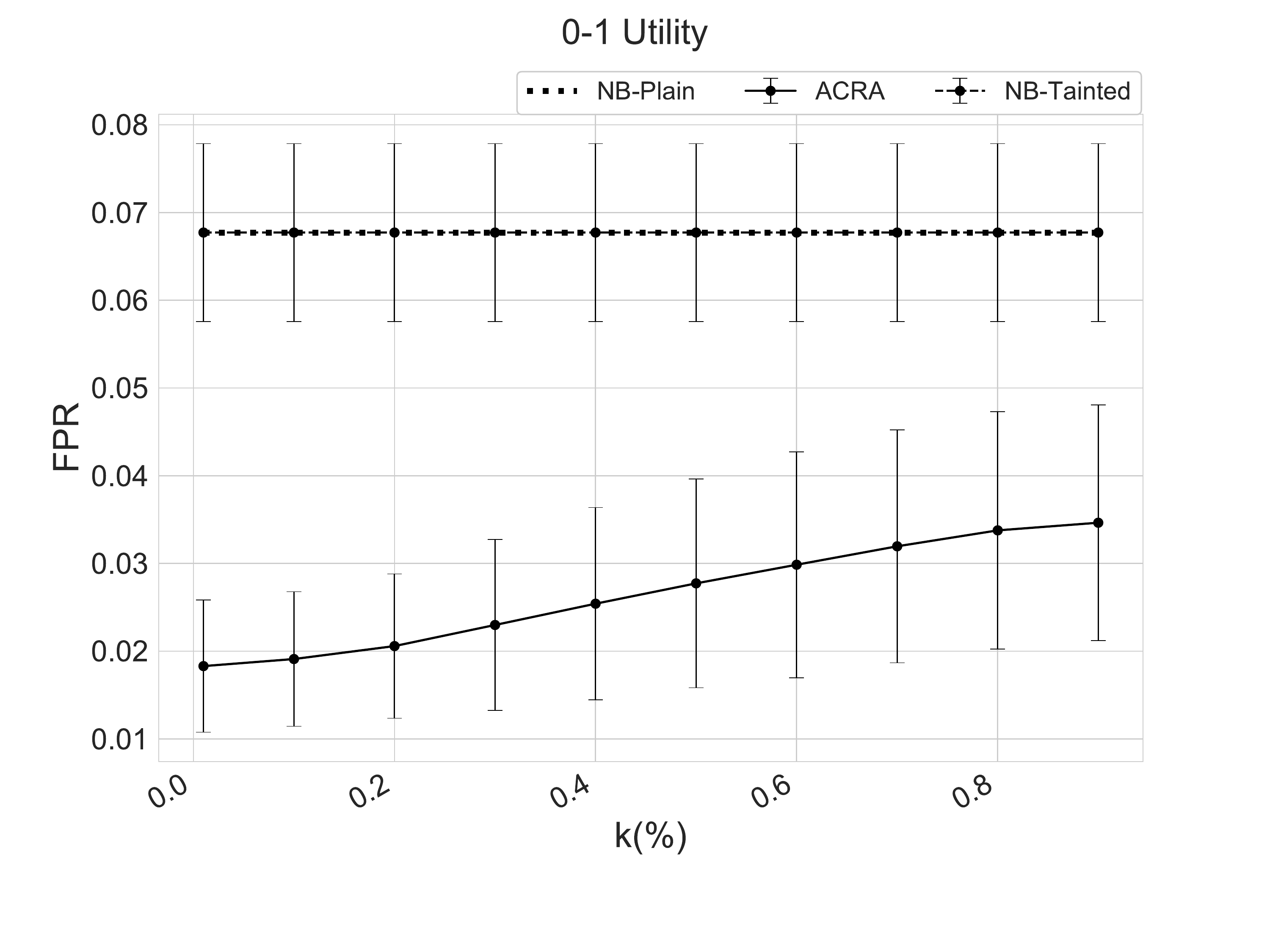}
    \end{minipage}%
    \begin{minipage}{0.5\textwidth}
        \centering
        \includegraphics[width=1\linewidth]{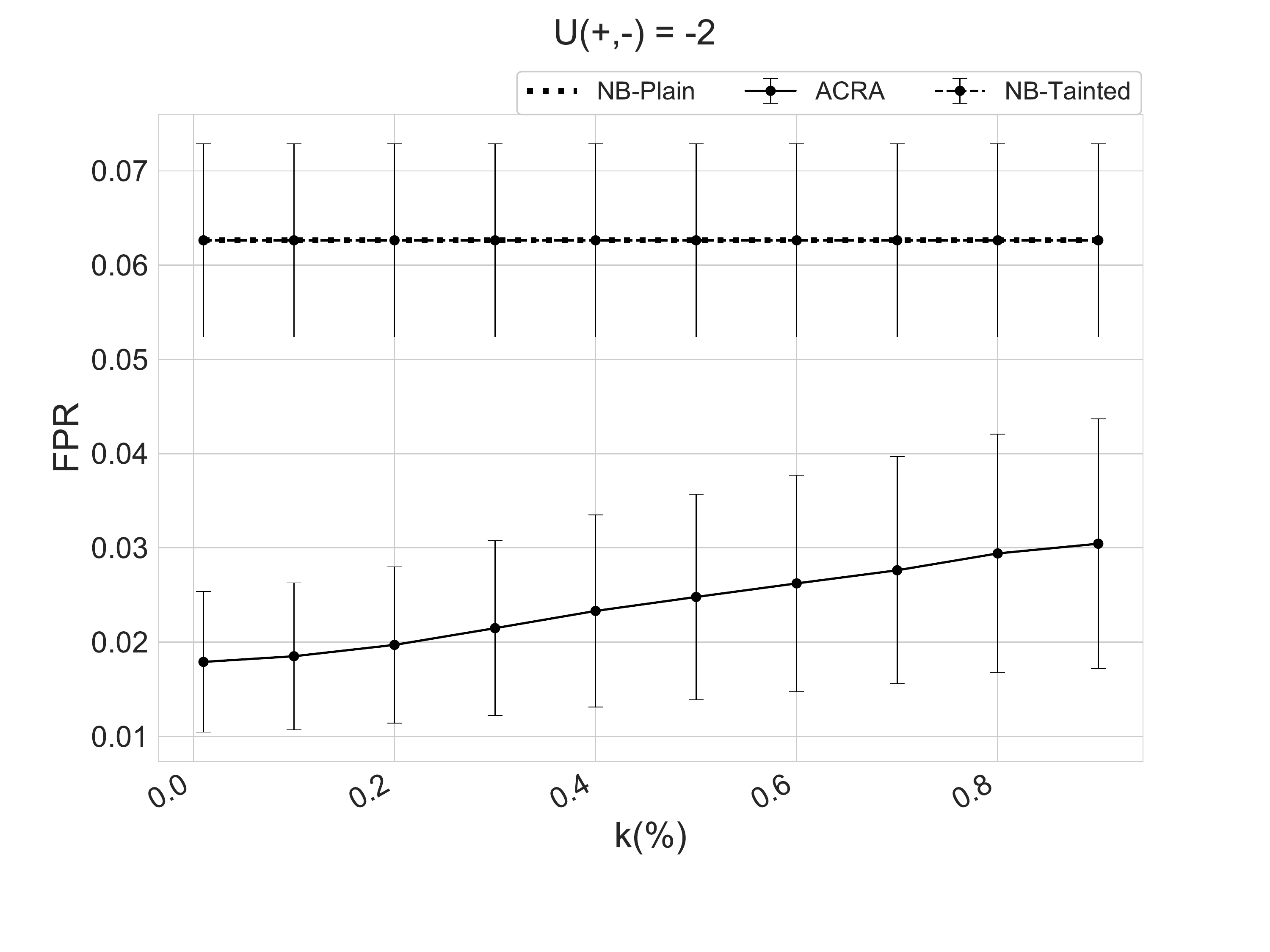}
    \end{minipage}
    \linebreak
    \begin{minipage}{.5\textwidth}
        \centering
        \includegraphics[width=1\linewidth]{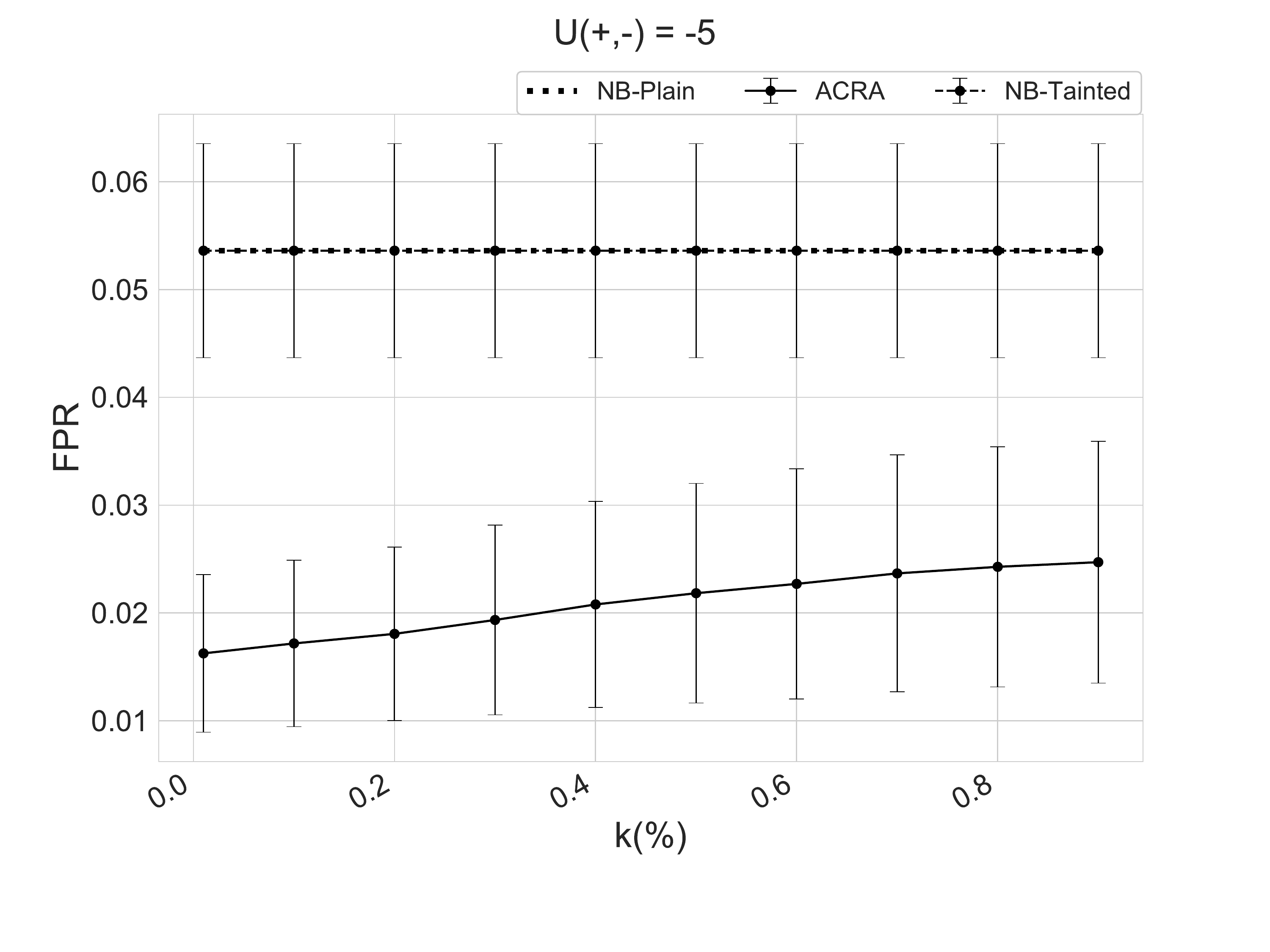}
    \end{minipage}%
    \begin{minipage}{0.5\textwidth}
        \centering
        \includegraphics[width=1\linewidth]{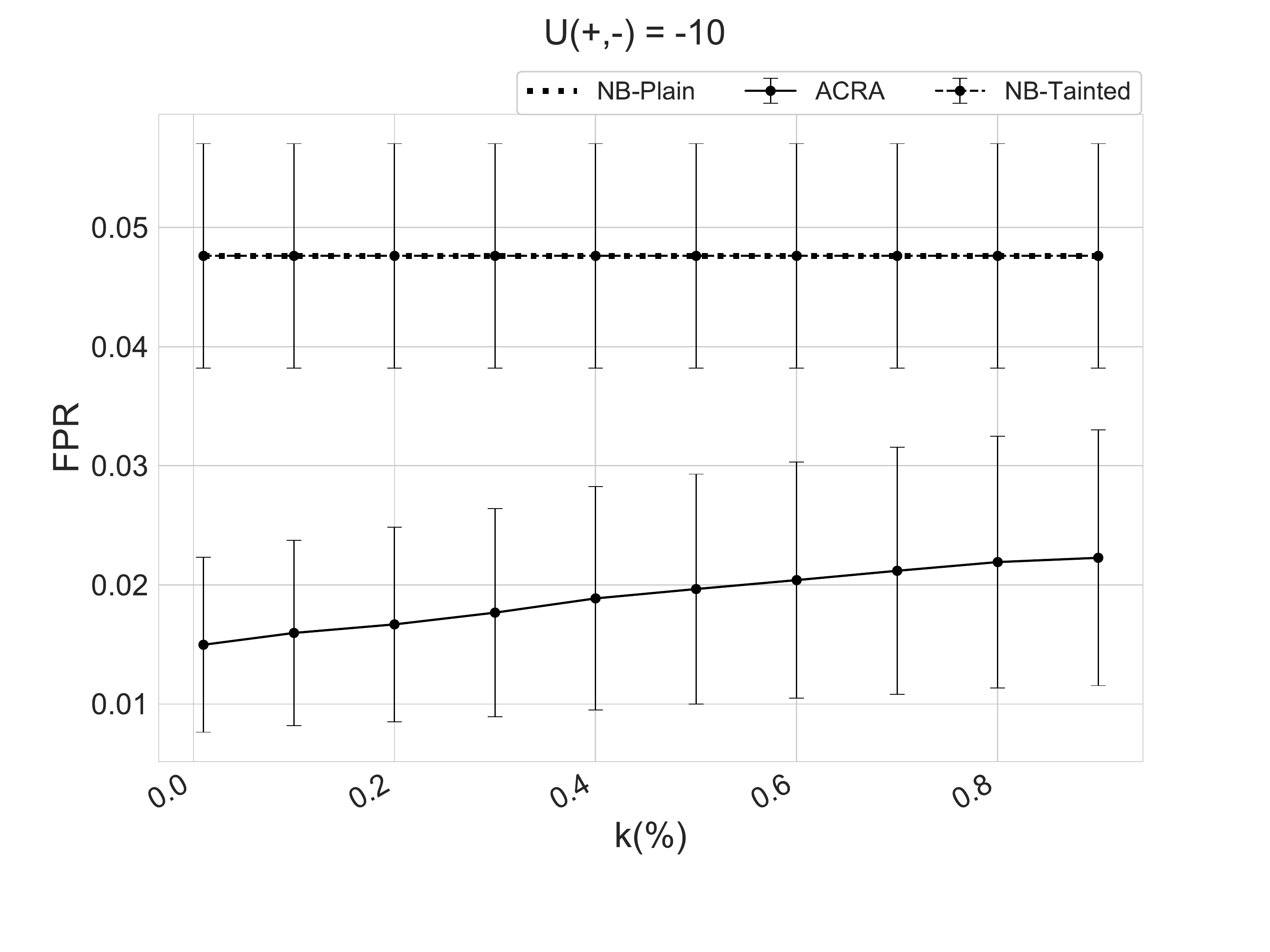}
    \end{minipage}
    \caption{Average false positive rate versus $k$ for different utility models.}\label{fput}
\end{figure}
\begin{figure}[ht]
    \centering
    \begin{minipage}{.5\textwidth}
        \centering
        \includegraphics[width=1\linewidth]{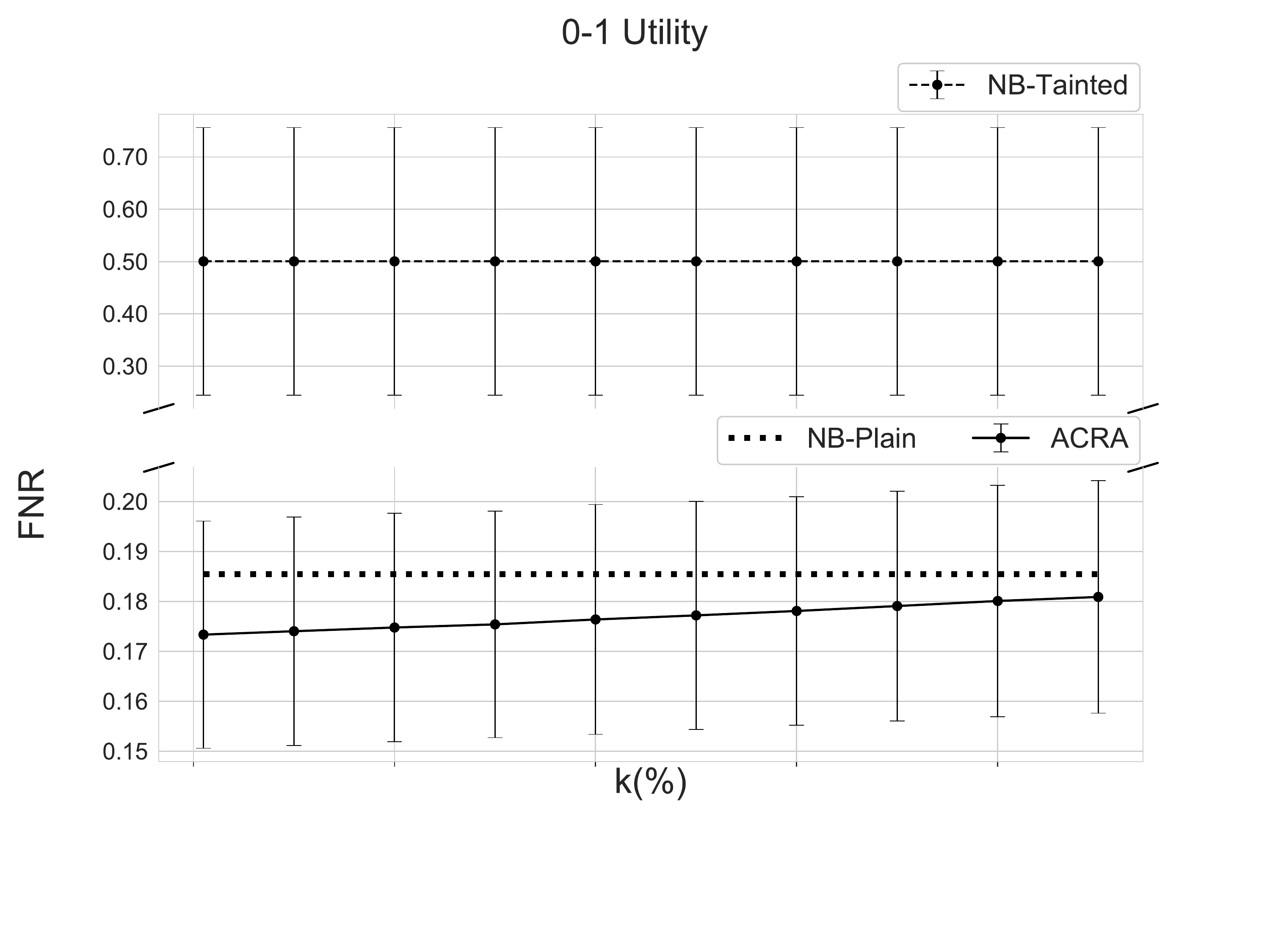}
    \end{minipage}%
    \begin{minipage}{0.5\textwidth}
        \centering
        \includegraphics[width=1\linewidth]{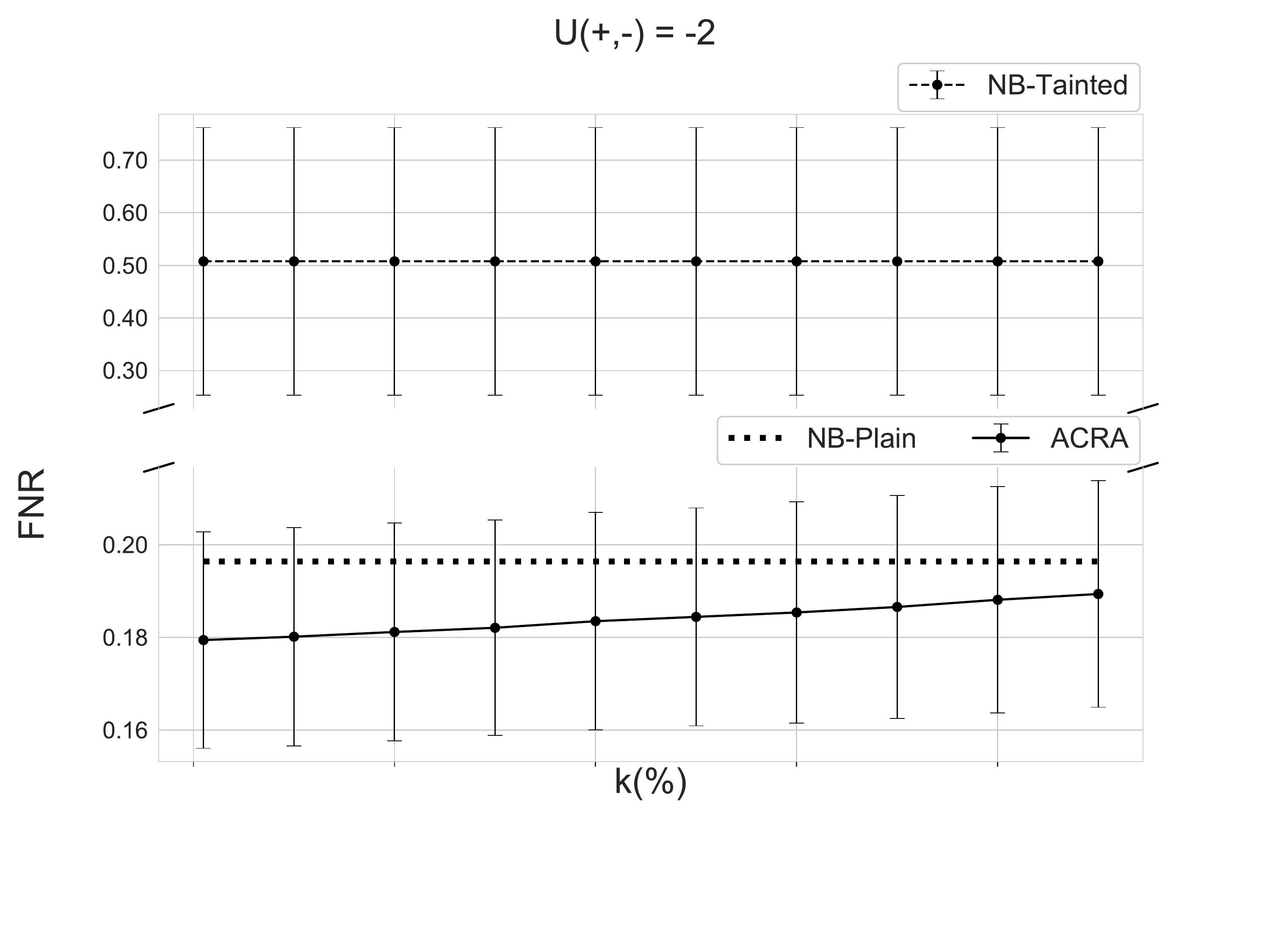}
    \end{minipage}
    \linebreak
    \begin{minipage}{.5\textwidth}
        \centering
        \includegraphics[width=1\linewidth]{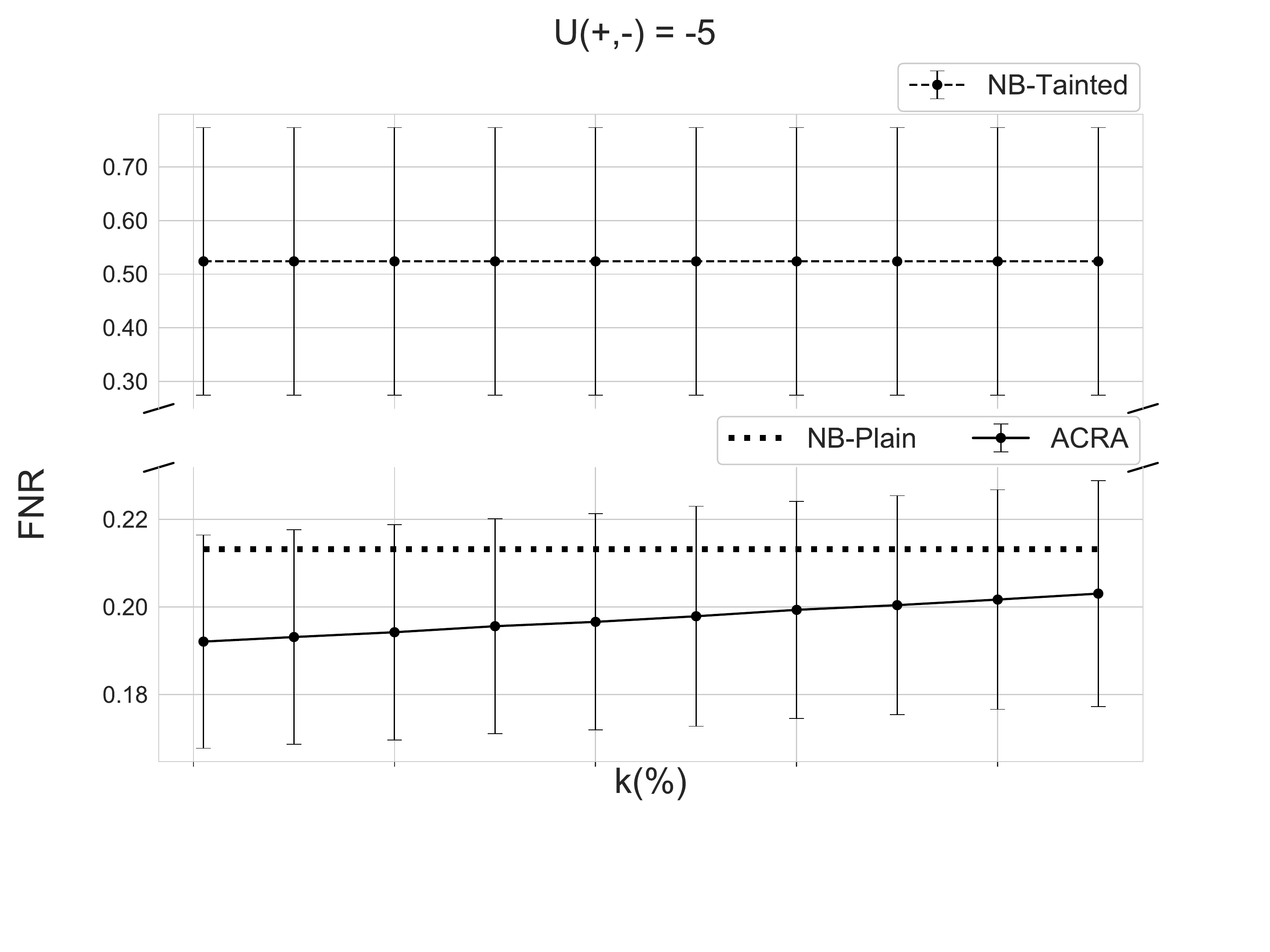}
    \end{minipage}%
    \begin{minipage}{0.5\textwidth}
        \centering
        \includegraphics[width=1\linewidth]{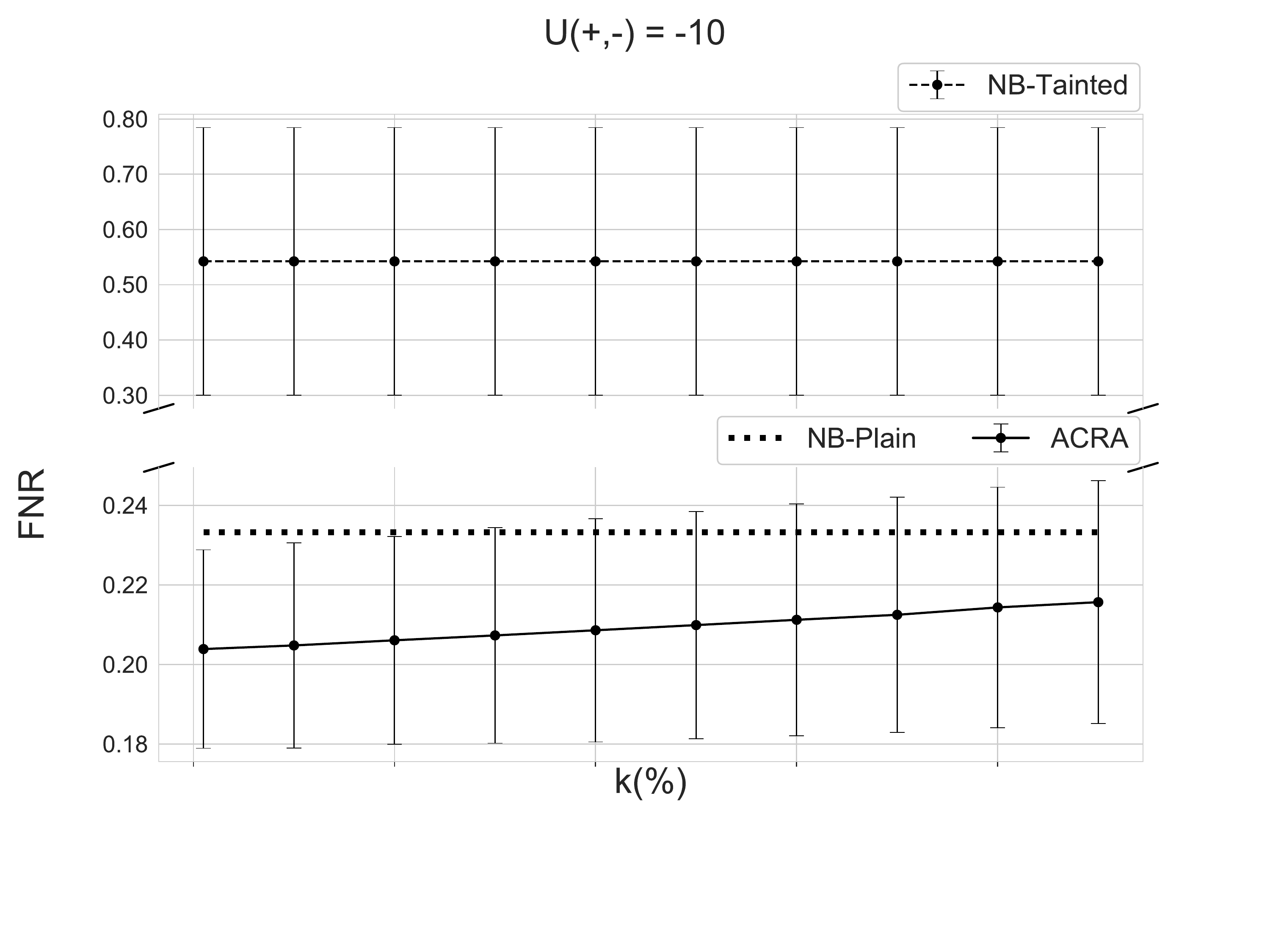}
    \end{minipage}
    \caption{Average false negative rate versus $k$ for different utility models. }\label{fnut}

\end{figure}

To better understand these results, we plotted FPR and FNR in Figures \ref{fput} and \ref{fnut}, respectively. FPR coincides for NB-Plain and NB-Tainted as the adversary is not modifying innocent instances. Both FPR and FNR grow with $k$ for ACRA: the more the classifier knows about the adversary strategy, the better she protects, and lower FPR and FNR are attained. In addition, an increase of $u_C(+,-)$ raises the cost of false positives, reducing FPR at the expense of increasing FNR.

Regarding the conceptual comparison of different algorithms, observe that false negatives undermine the performance of NB on tampered data. In contrast, ACRA seems more robust, presenting smaller FNR than NB-Tainted. ACRA has also significantly lower FPR than NB, causing the overall performance to raise up.  The reason for this is that the adversary is very unlikely to apply the identity attack to a spam, as the cost difference between such attack and 1-GWI attacks is small in terms of utility gain. Then, for a legitimate email that NB classifies as positive, i.e.\ has high $p_C(x|+)$, ACRA will give a very low weight to $p_C(x|+)$, thus reducing the probability of classifying such email as spam. Reducing FPR is crucial in spam detection, as filtering out a non-spam is typically more undesirable than letting spam reach the user.
Interestingly enough, ACRA also has lower FNR than NB-Plain, specially for low values of $k$, although this improvement is not as remarkable as that for FPR. Both effects together produce ACRA to outperform utility sensitive NB both in tainted and untainted data: we enhance the classifier performance by taking advantage of the information we may have about the adversary, a core idea underlying ACRA.
\subsection{Robustness}
We have tested the ACRA algorithm against an adversary whose parameters were fixed to the expected values of the distributions assumed by the classifier. It is natural to ask how departures from such assumptions about the adversary's behaviour would affect performance. In this section, we compare the robustness of ACRA and the corresponding game theoretic solution, to departures from the assumed value of $p_{a(x)}^A$, the expected value of the probability that $A$ concedes to $C$ saying that the instance $a(x)$ is malicious. As we discussed above, this is clearly the most challenging assessment. 

\begin{figure}[hbt]
\subfloat[\label{robust1}]
{\includegraphics[width=0.5\linewidth]{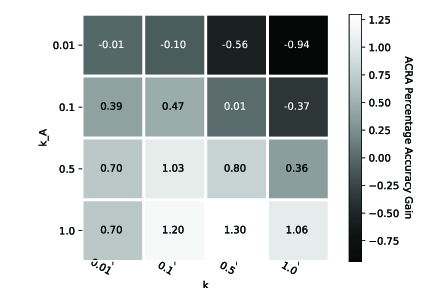}}
\subfloat[\label{robust2}]{\includegraphics[width=0.5\linewidth]{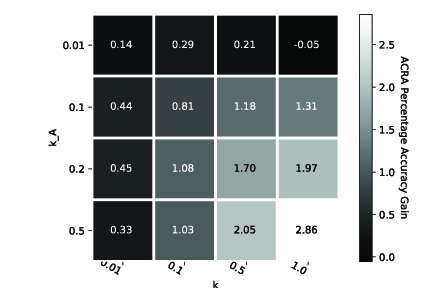}}
\caption{ACRA accuracy gain with respect to the game theoretic solution.} \label{robust}
\end{figure}

The game theoretic solution assumes common knowledge. For the case of $p_{a(x)}^A$ this means that this quantity is fixed at a certain value, shared by both $A$ and $C$. Thus $C$, when modelling $A$, assumes that he is using a certain value of $p_{a(x)}^A$ that coincides with the one that he is actually using. For robustness purposes, we attacked the test set solving the attacker problem \eqref{attackerprob} for each test instance $x$, but using this time perturbed values of $p_{a(x)}^A$ around the one assumed by $C$. To do so, we sample $p_{a(x)}^A$ from a beta distribution centered at the assumed value, with variance $k_A\Delta$, where $k_A$ is the proportion of the maximum allowed variance $\Delta$ (again, we require the density to be concave in its support). We compare performance against this attacker of both the ACRA solution, which models uncertainty on $p_{a(x)}^A$ by placing a beta distribution centered at the assumed value of $p_{a(x)}^A$ and variance $k\Delta$, and the game theoretic solution for which $p_{a(x)}^A$ is fixed at its assumed value. We performed experiments for different values of $k$ and $k_A$. Results for the percentage accuracy gain of ACRA, i.e.\ ACRA percentage accuracy minus game theory percentage accuracy, are presented in Figure \ref{robust1}. As can be seen, if the attacker behaves closely to the common knowledge assumptions (i.e.\ low $k_A$), then the game theoretic solution and ACRA with low variance behave similarly. If in this case, we increase the variance assumed by ACRA, then its performance degrades, as it is overestimating the uncertainty. If the variance of the distribution that the adversary puts in $p_{a(x)}^A$ is high (i.e.\ high $k_A$), then big deviations from the common knowledge value of $p_{a(x)}^A$ are more likely, thus degrading the performance of the game theoretic solution. Nevertheless, in this case ACRA remains robust to these perturbations, thanks to accounting for uncertainty on the adversary's value of $p_{a(x)}^A$. In addition, although not shown in the Figure, ACRA's accuracy was above 0.89 for every pair $(k, k_A)$, beating always the NB algorithm.

One could argue that this comparison is not sufficiently fair, as we are sampling the adversary's values of $p_{a(x)}^A$ from a beta distribution, that is the one assumed by ACRA. Figure \ref{robust2} shows the results of an alternative experiment. In this case, to perturb the values of $p_{a(x)}^A$ of each test instance, we add a number, uniformly distributed in the interval $[-k_A, k_A]$, to the common knowledge value. As can be seen, in this case ACRA is again more robust to this imprecision in the attacker's model, thus providing better generalization. In addition, ACRA's accuracy was not excessively damaged despite using perturbed values of $p_{a(x)}^A$, being above 0.87 for every pair $(k, k_A)$.


\section{Computational issues}
\label{sec:comp}

The raw version of ACRA presented above may turn out to be extremely heavy computationally in some application domains in which little assumptions about the adversary behaviour are made. We first asses ACRA computationally and then propose several solutions.

\subsection{Computational assessment}

Note first that if no assumptions about the attacks are made, the sets $\mathcal{A}(x)$, and consequently $\mathcal{X}'$, could grow rapidly. The size of these sets strongly depends on the application domain. For instance, in spam detection the number of possible adversarial manipulations is $2^n$, where $n$ is the number of words considered by $C$ to undertake the classification. 

In fact, the size of $\mathcal{X}'$ affects critically the number of computations. Notice that in order to classify a given instance $x'$, we need to estimate $p_C(a_{x \rightarrow x'} |x,+)$ for each $x \in \mathcal{X}'$ to compute the summation in \eqref{pis}. In addition, estimating each $p_C(a_{x \rightarrow x'} |x,+)$ requires a MC simulation with size $K$, see \textsc{Estimate} $p_{C}(a_{x \rightarrow x'} |x,+)$ routine of the Appendix. Thus, if $G$ represents the computational cost of each MC sample, the overall cost to classify one instance would be $K \cdot |\mathcal{X}'| \cdot G$. Moreover, $G$ could depend on the size of $\mathcal{A}(x)$. 

We study now how to reduce the computational burden of ACRA by means of reducing the size of $\mathcal{X}'$ and the MC size $K$ without affecting performance too much.



\subsection{Computational enhancements}

One possibility to reduce $|\mathcal{X}'|$ could be to use application-specific information to restrict the class of possible attacks. For instance, if we consider only $k$-GWI attacks in spam detection, the size of $\mathcal{X}'$ is reduced to $\mathcal{O}(n^k)$. In addition, case-specific constraints about the adversary behavior could be used to achieve a greater decrease. For example, in GWI we may reduce $n$ assuming that some words cannot be modified by the adversary, or limit the number of words inserted, either explicitly, or implicitly through penalizing the insertion of additional words.

Apart from using application-specific information, we present now several general suggestions which can be used to alleviate the computational burden. Note first that the optimization problem \eqref{pis} may be reformulated as setting $c(x') = +$ if and only if $\sum_{x \in \mathcal{X}'} p_C(a_{x \rightarrow x'} |x,+) p_C(x|+) > t$, where
\begin{eqnarray*}
t = \frac{\bigg[u_C(-, -) - u_C(+, -)\bigg] p_C(x'|-)p_C(-)}{\bigg[ u_C(+, +) -  u_C(-, +)\bigg] p_C(+)  }.
\end{eqnarray*}
Rather than going through the whole $\mathcal{X}'$, we can approximate the left hand side summation through Monte Carlo. Should $\{ x_n \}$ be a sample of size $N$ from $p_C (x|+)$, with $N$ lower than the cardinality of $\mathcal{X}'$, the condition would be
approximated through
\begin{eqnarray}
I = \frac{1}{N} \sum_{n=1}^N p_C(a_{x_n \rightarrow x'} |x_n ,+) I (x_n \in \mathcal{X}') >   t. \label{threshold1}
\end{eqnarray}
A potential problem with this approach is that $p_C (x|+)$ for $x \in \mathcal{X}'$ is generally small and a standard MC sample might contain few points in $\mathcal{X}'$. We could use 
importance sampling, \cite{owen2000}, to mitigate this issue for example using the restriction of $p(x|+)$ to $\mathcal{X}'$ as importance distribution. Let $\tilde{p}(x|+)$ be the probability distribution defined by
\begin{eqnarray*}
\tilde{p}(x|+) = \frac{p(x|+)}{Q} \cdot I(x \in \mathcal{X}')
\end{eqnarray*}
with $Q = \sum_{x \in \mathcal{X}'} p(x|+)$ . Then 
\begin{eqnarray*}
\sum_{x \in \mathcal{X}'} p_C(a_{x \rightarrow x'} |x,+) p_C(x|+) &=& \sum_{x \in \mathcal{X}'} \frac{p_C(a_{x \rightarrow x'} |x,+) p_C(x|+)}{\tilde{p}(x|+)} \tilde{p}(x|+) \\
&=& Q \sum_{x \in \mathcal{X}'} p_C(a_{x \rightarrow x'} |x,+) \tilde{p}(x|+).
\end{eqnarray*}
Now, if $\{ x_n \}$ is a sample of size $N$ from $\tilde{p}(x|+)$, we could approximate our target quantity $I$ by
\begin{eqnarray}
\tilde{I} = \frac{Q}{N} \sum_{n=1}^N p_C(a_{x_n \rightarrow x'} |x_n,+). \label{II}
\end{eqnarray}

We should take into account, though, the inherent uncertainty in the above MC approximations when checking inequality \eqref{threshold1}. 
An estimate $\tilde{\Delta}$ of its standard deviation would be
$$
\tilde{\Delta} = \sqrt{\frac{\sum_{i=1}^N Q^2 p_C^2(a_{x_i \rightarrow x'} |x_i ,+) - N\tilde{I}^2 }{ N-1 }}.
$$
We would then declare $c(x') = +$ if $\tilde{I} - 2 \tilde{\Delta}  >  t$. 
Observe that we could test this condition sequentially, before reaching the maximum size $N$ allowed by our computational budget. By making $\tilde{I}_m$ and $\tilde{\Delta}_m$ depend on the sample size $m$, we would check sequentially whether
\begin{eqnarray}
\tilde{I}_m - 2 \tilde{\Delta}_m  >  t, \label{seq}
\end{eqnarray}
and stop if verified, with $\tilde{I}_m$ and $\tilde{\Delta}_m$ defined sequentially, since 
\begin{eqnarray*}
\tilde{I}_m &=& \frac{ (m-1) \tilde{I}_{m-1}  + p_C(a_{x_m \rightarrow x'} | x_m, +)}{m}, \\ \tilde{\Delta}_m^2 &=& \frac{(m-2)\tilde{\Delta}_{m-1}^2 + Q^2p_C^2(a_{x_m \rightarrow x'} | x_m, +) + (m-1)\tilde{I}_{m-1}^2 - m\tilde{I}^2_m}{m-1}.
\end{eqnarray*}
%
Note that with the proposed enhancement, the number of computations is $K \cdot N$ rather than $K \cdot |\mathcal{X}'|$ , where $N$ is the chosen sample size. Thus, we manage to eliminate the dependence on $|\mathcal{X}'|$. $N$ should be fixed to trade off between accuracy and speed.

A complementary possibility to speed up computations is to use a relatively small MC size $K$ for estimating $p_{C}(a_{x \rightarrow x'} |x,+)$. To mitigate getting null probabilities, we could adopt a Dirichlet-multinomial model with non-informative prior $Dir (1,1,...,1)$ over the probabilities of the attacks in $\mathcal{A}(x)$ and approximate such probability through
\begin{eqnarray*}
\widehat{p}_C ( a_{x \rightarrow x'}\,|\,x, +) = \frac{ \# \{a_k^* =  a_{x \rightarrow x'}\}+ 1}{K + |\mathcal{A}(x)|}.
\end{eqnarray*}
Moreover, we could use a regression metamodel, \cite{kleijnen1992regression}, based on approximating in detail $p_C(a_{x \rightarrow x'} |x,+)$ at some pairs $(x,x')$, as allowed by our computational budget, fit a regression model $\psi (x,x')$ to $(x, x',$ ${\widehat{p}_C} (a_{x \rightarrow x'} |x,+))$ and use it to replace $p_C(a_{x \rightarrow x'} |x,+)$ in the above expressions.

Last, but not least, ACRA is amenable of parallelization in at least two respects. Observe first that the terms in summations \eqref{threshold1} or \eqref{II} may be evaluated independently. To improve performance, we may compute batches of those terms in parallel by running different processes at different nodes of a multi-core cluster. Whenever we use the sequential approach in \eqref{seq}, this parallelization strategy would require a master node that checks such condition periodically when the computation of any batch of  terms finishes in the corresponding worker node. Finally, recall that computing terms in summations (6) or \eqref{II} entails a simulation. We could accelerate the computation of each simulation by running in parallel different processes for different batches of MC samples. Both parallelization strategies could be combined by sending different simulations to different nodes in the cluster, and parallelizing each simulation within the cores of each node.




The combination of the above approaches alleviates tremendously the computational burden and
largely makes ACRA computationally feasible as we show next.
\section{Application} \label{sec:Application}
We illustrate several of the proposed enhancements with the example in Section \ref{sec:example}. We start by testing the ACRA framework using MC simulation (MC ACRA) with importance sampling and the sequential strategy \eqref{seq} against the raw algorithm (ACRA). We tried different MC sample sizes $N$, measuring them as proportions of the cardinal of $\mathcal{X}'$: e.g.\ an MC sample size of 0.5 corresponds to considering $N = |\mathcal{X}'|/2$ values. We fixed $K=1000$, the MC size in the \textsc{Estimate} $p_{C}(a_{x \rightarrow x'} |x,+)$ function, the adjustable var parameter $k$ to $0.1$, and used the $0/1$ utility.
\begin{table}[hbt]
\centering
\begin{spacing}{0.2}
\begin{tabular}{ccccc}
\toprule

                          & \textbf{Size}      & \textbf{Accuracy}     & \textbf{FPR} & \textbf{FNR}        \\ \midrule

\textbf{ACRA}                      & 1.00      & $0.919 \pm 0.010$      & $0.019 \pm 0.008$	& $0.177 \pm 0.022$            \\ \midrule

\textbf{MC ACRA}                   & 0.75  & $0.912 \pm 0.012$      & $0.032 \pm 0.009$    & $0.174 \pm 0.023$                             \\ \midrule

\textbf{MC ACRA}                   & 0.50  & $0.905 \pm 0.016$      & $0.027 \pm 0.009$    & $0.199 \pm 0.032$                             \\ \midrule

\textbf{MC ACRA}                   & 0.25  & $0.885 \pm 0.029$      & $0.021 \pm 0.007$    & $0.260 \pm 0.067$              \\ \midrule

\textbf{MC ACRA}                   & 0.10  & $0.841 \pm 0.047$      & $0.016 \pm 0.005$    & $0.370 \pm 0.120$              \\ \midrule

\textbf{NB-Tainted}                      & -      & $0.761 \pm 0.101$      & $0.680 \pm 0.100$	& $0.500 \pm 0.250 $ 

\\ \bottomrule
\end{tabular}
\end{spacing}
\vspace{0.4cm}
\caption{Comparison between MC ACRA, raw ACRA and NB.}\label{tab:comp}
\end{table}
Table \ref{tab:comp} shows the average performance metrics, with standard deviations, of the algorithms over 100 experiments. Note that as the sample size $N$ increases, accuracy also increases. Nevertheless, we get fairly good results for relatively small sample sizes. For example, with just a 0.1 sample size we manage to beat NB in accuracy as well as in FPR and FNR.  Considering a 0.5 sample size, we almost recover the original performance levels. 

\begin{figure}[hbt]
\subfloat[\label{su1}]
{\includegraphics[width=0.5\linewidth]{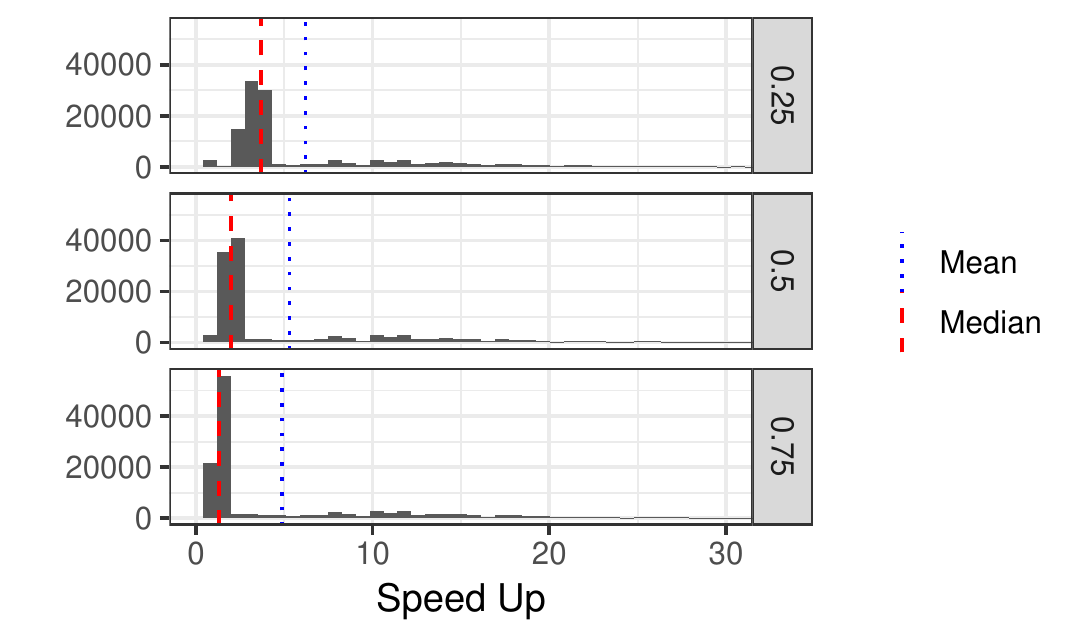}}
\subfloat[\label{su2}]{\includegraphics[width=0.5\linewidth]{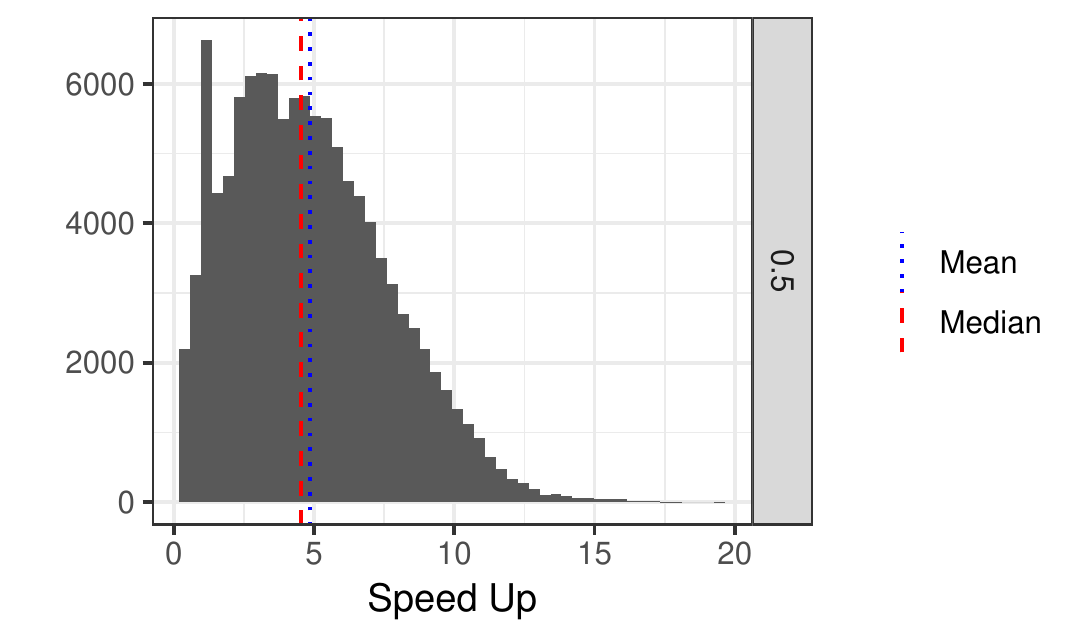}}
\caption{Speed up histograms.} \label{su}
\end{figure}

To compare execution times, we computed the speed up (quotient between execution times of ACRA and MC ACRA over all 100 experiments). Figure \ref{su}a presents the speed up histogram, Table \ref{tab:speedup} shows mean and median speed ups for the MC sizes (0.25, 0.5, 0.75).
As expected, the median is close to the inverse of the MC size, e.g.\ when size is 0.5, MC ACRA performs approximately twice faster than ACRA. Nevertheless, the speed up distributions (Figure \ref{su1}) are skewed to the right suggesting that MC ACRA performs much faster on average. This is due to the sequential rule \eqref{seq}: for some instances, such condition is reached in a few iterations and, consequently, over those instances MC ACRA performs much faster than ACRA.
\begin{table}[hbt]
\centering
\begin{spacing}{0.2}
\begin{tabular}{ccc}
\toprule \\
\textbf{Size}      & \textbf{Mean}     & \textbf{Median}  \\ \midrule
0.25               &   6.20                     &  3.69                       \\ \midrule
0.50               &   5.30                     &  2.00                       \\ \midrule
0.75               &   4.86                     &  1.31                       \\
\bottomrule
\end{tabular}
\end{spacing}
\vspace{0.4cm}
\caption{Mean and median speed ups.}\label{tab:speedup}
\end{table}

We have also tested the first parallelization approach in Section \ref{sec:comp}, computing in parallel the terms in the MC approximation \eqref{II}. We used a 16 core processor for this purpose. We performed 100 experiments fixing the variance parameter $k = 0.1 $, MC size to 0.5 and $0/1$ utility. The histogram of speed ups is in Figure \ref{su2}. In this case, both the mean (4.856) and median (4.530) are close. We do not use the sequential approach \eqref{seq} and consequently, extreme values do not occur. Nevertheless, we obtain a huge improvement in time performance, almost 5 times faster both in mean and median.

The combination of the above approaches induces considerable improvements rendering ACRA largely feasible, as we illustrate with the results of an experiment under 2-GWI attacks with different databases\footnote{Besides the UCI Spam Data Set, we used the Enron-Spam Data Set at \url{https://www.cs.cmu.edu/~enron} and the Ling-Spam Data Set at \url{http://csmining.org/index.php/ling-spam-datasets.html}} in Table \ref{tab:comp2gwi}. As in previous examples, we report averages over 100 experiments performed under different train-test splits. Observe that MC ACRA with size 0.5 consistently beats utility sensitive NB.

\begin{table}[h!]
\centering
\small
\begin{spacing}{0.2}
\begin{tabular}{cccccc}
\toprule

                          &\textbf{Dataset} &  \textbf{Accuracy}     & \textbf{FPR} & \textbf{FNR}        \\ \midrule


\textbf{MC 0.5 ACRA}        &UCI           &  $0.904 \pm 0.012$      & $0.037 \pm 0.007$    & $0.187 \pm 0.023$                             \\ \midrule

\textbf{NB-Tainted}       &UCI                   & $0.724 \pm 0.088$     & $0.066 \pm 0.008$	& $0.601 \pm 0.022$            \\ \midrule


\textbf{MC 0.5 ACRA}        &Enron-Spam            & $0.824 \pm 0.017$      & $0.132 \pm 0.012$    & $0.305 \pm 0.073$                             \\ \midrule

\textbf{NB-Tainted}       &Enron-Spam                     & $0.534 \pm 0.011$     & $0.283 \pm 0.013$	& $1.000 \pm 0.000$            \\  \midrule


\textbf{MC 0.5 ACRA}        &Ling-Spam            & $0.958 \pm 0.008$      & $0.039 \pm 0.001$    & $0.057 \pm 0.030$                             \\ \midrule

\textbf{NB-Tainted}       &Ling-Spam                    & $0.800 \pm 0.016$     & $0.040 \pm 0.001$	& $1.000 \pm 0.000$          


 \\ \bottomrule
\end{tabular}
\end{spacing}
\vspace{0.4cm}
\caption{Comparison between size 0.5 MC ACRA and NB under 2-GWI attacks.}\label{tab:comp2gwi}
\end{table}
\section{Discussion}
Adversarial classification aims at enhancing classification algorithms to achieve robustness in presence of adversarial examples, as usually encountered in many security applications. The pioneering work of \cite{adversarialClassification2004} framed most of later approaches to adversarial classification within the standard game theoretic paradigm, in spite of the unrealistic common knowledge assumptions required, actually even questioned by the authors. This motivated us to focus on an ARA perspective to the problem presenting ACRA, a general framework for adversarial classification that mitigates such assumption. Our framework is general in the sense that application-specific assumptions are kept to a minimum. Also, we have provided empirical evidence supporting the robustness of the ACRA framework to imprecisions in the assumptions made about the adversary. In particular, ACRA has been shown to be more robust than its common knowledge, purely game theoretic counterpart. Finally, we have presented computational enhancements that have significantly improved ACRA performance, allowing us to solve large problems and use it in operational settings.

Our framework may be extended in several ways. First of all, in our examples we have used NB as the basic classifier in the preprocessing phase. We could use other generative classifiers, such as variational autoencoders, or even a mixture of them. It would be very interesting to extend the framework to use it with discriminative classifiers, specifically based on neural networks.

We could extend the ACRA approach to situations in which there is repeated play of the adversarial classification game, thus introducing the possibility of learning the adversarial utilities and probabilities in a Bayesian way.  

We have only considered exploratory attacks, but we could extend the approach to take into account attacks over the training data, called poisoning attacks, \cite{biggio2012poisoning}. In addition, we have just considered the case in which the attacker performs intentional attacks. In some problems, there could be, in addition, random attacks. The proposed framework could be adapted to take those into account as well as to the case in which there are several attackers. 

Finally, our work could be extended to the case of attacks to innocent instances (not just integrity violation ones). In this case, when computing $p_C(x'|-)$ in \eqref{pis} we must proceed similarly as we did when computing $p_C(x'|+)$: we consider all possible originating instances $x$ leading to $x'$, and sum them weighting each of them with their $p_C(a_{x \rightarrow x'}| x, -)$, the probability of the attacker choosing the attack linking each of them with $x'$, given that they are innocent. This would just involve replacing $p_C(x'|-)$ by $\sum_{x \in \mathcal{X}'} p_C(a_{x \rightarrow x'}| x, -)p_C(x|-)$ in \eqref{pis}. Finally, as computing both $p_C(a_{x \rightarrow x'}| x, -)$ and $p_C(a_{x \rightarrow x'}| x, +)$ demands strategic thinking, we need to consider the attacker's problem as we did in Section \ref{subsec:theAdversaryProblem}, but this time allowing the attacker to modify also innocent instances.

We have concentrated on binary classification problems, but the extension to multi-label classification is relevant. This would entail including the summands corresponding to each class in $\eqref{pis}$ and building an appropriate attacker model depending on his particular interests: for instance, he could be interested on making the classifier mislabel any instance or in making her classify instances within a particular group of classes. It is clear that the presence of an adversary would invalidate the one-vs-one and one-vs-rest approaches to multiclass classification, as this procedure could be easily exploited. This would affect the choice of the base classifier within the ACRA approach.

We have illustrated the approach in spam detection, but other security areas like malware or fraud detection are truly important. Finally, note that in ACRA we go through a simulation stage to forecast attacks and an optimization stage to determine optimal classification. The whole process might be performed in a single stage, possibly based on augmented probability simulation, \cite{bielza1999decision}.

\section*{Acknowledgements}
R.N. acknowledges support the Spanish Ministry for his grant FPU15-03636. The work of D.R.I. is supported by the Spanish Ministry program MTM2017-86875-C3-1-R and the AXA-ICMAT Chair on Adversarial Risk Analysis. This work has also been partially supported by the Spanish Ministry of Economy through the Severo Ochoa Program for Centers of Excellence in R\&D (SEV-2015-0554), the project MTM2015-72907-EXP and the EU's Horizon 2020 project 740920 CYBECO (Supporting Cyberinsurance from a Behavioural Choice Perspective). F.R. acknowledges the contribution of the \textit{Comunidad de Madrid} through its Chair of Excellence programme. We are grateful for the suggestions of the referees.

\section*{References}

\bibliography{referencesA}

\pagebreak
\section*{Appendix}
We include here the routines to generate from the random utility function (Routine 1) and estimate $p_{C}(a_{x \rightarrow x'} |x,+)$ (Routine 2) required in the spam detection problem.

\subsection*{Routine 1}
\begin{algorithmic}
\Function{Generate $U^k(y_{C}, +, a)$}{}
\State Generate $B^k \sim B$, $\rho^k \sim U [a_1,a_2]$
\State Generate  $Y_{++}^k \sim -Ga (\alpha _1, \beta _1) $, $ Y_{-+}^k \sim Ga (\alpha _2 , \beta _2 )   $
\State $U^k(y_{C}, y, a) = \exp ( \rho^k (Y_{y_C , y} ^k - B^k ) )$
\State \Return $U^k(y_{C}, y, a)$
\EndFunction
\end{algorithmic}

\noindent where $k$ designates a generic sample instance in the Monte Carlo scheme.
\subsection*{Routine 2}

\begin{algorithmic}
\Function{Estimate $p_{C}(a_{x \rightarrow x'} |x,+)$}{}
\For{$x \in \mathcal{X}'$}
\State Compute $\mathcal{A}(x)$
\For{$a \in \mathcal{A}(x)$}
	\State Compute $a(x)$ and $r_{a}$ using \eqref{ra}
	\State Using $r_a$ and $var$, compute $\delta_{1}^{a}$, $\delta_{2}^{a}$ as in \eqref{deltas}
\EndFor

\For{$k = 1,2,\dots, K$}

\For{$a \in \mathcal{A}(x)$}
	\State Generate $U^k(y_{C}, +, a)$,  $P^{Ak} _{a  } \sim \beta e(\delta_{1}^{a},\delta_{2}^{a})$
	\State Compute $\psi ^k (a) =    \left[   U_A^k( +, +, a) - U_A^k( -, +, a) \right] P_{a}^{Ak}   +    U_A^k( -, +, a) $
\EndFor
\State Compute $ a_k^* = \argmax _{a\in \mathcal{A}(x) } \psi ^k (a) $
\EndFor
\State Compute
 $$
  \widehat{p}_C ( a_{x \rightarrow x'}\,|\,x, +) = \frac{ \# \{a_k^* =  a_{x \rightarrow x'}\}}{K }
$$
\State Store $\widehat{p}_C ( a_{x \rightarrow x'}\,|\,x, +)$
\EndFor
\EndFunction
\end{algorithmic}

\end{document}